\documentclass{article}
\usepackage{microtype}
\usepackage{graphicx}
\usepackage{subfigure}
\usepackage{wrapfig}
\usepackage{arydshln}
\usepackage{booktabs} 
\usepackage{hyperref}

\usepackage[accepted]{icml2025}

\usepackage{amsmath}
\usepackage{amssymb}
\usepackage{mathtools}
\usepackage{amsthm}

\usepackage{arydshln}
\usepackage{multirow}
\usepackage{multicol} 
\usepackage{makecell}
\usepackage{etoolbox}
\makeatletter
\usepackage{wrapfig}
\usepackage{caption}
\usepackage{microtype}
\usepackage{graphicx}
\usepackage{booktabs}
\usepackage{ulem}
\usepackage{amsmath, amsthm, amssymb}
\usepackage{bm}
\usepackage{pifont}

\usepackage[capitalize,noabbrev]{cleveref}

\newtheorem{definition}{Definition}
\newtheorem{theorem}{Theorem}

\newtheorem{proposition}{Proposition}

\newtheorem{corollary}{Corollary}

\newtheorem{assumption}{Assumption}

\newtheorem{innercustomgeneric}{\customgenericname}
\providecommand{\customgenericname}{}
\newcommand{\newcustomtheorem}[2]{%
  \newenvironment{#1}[1]
  {%
   \renewcommand\customgenericname{#2}%
   \renewcommand\theinnercustomgeneric{##1}%
   \innercustomgeneric
  }
  {\endinnercustomgeneric}
}
\newcustomtheorem{customthm}{Theorem}
\newcustomtheorem{customcor}{Corollary}
\newcustomtheorem{customprop}{Proposition}
\newcustomtheorem{customassum}{Assumption}

\usepackage{color}

\theoremstyle{plain}
\theoremstyle{definition}
\theoremstyle{remark}

\usepackage[disable,textsize=tiny]{todonotes}

\begin{document}

\twocolumn[
\icmltitle{Model Reprogramming Demystified:  A Neural Tangent Kernel Perspective}



\icmlsetsymbol{equal}{*}

\begin{icmlauthorlist}
\icmlauthor{Ming-Yu Chung}{duke,ntu}
\icmlauthor{Jiashuo Fan}{duke}
\icmlauthor{Hancheng Ye}{duke}
\icmlauthor{Qinsi Wang}{duke}
\icmlauthor{Wei-Chen Shen}{ntu}
\icmlauthor{Chia-Mu Yu}{nycu}
\icmlauthor{Pin-Yu Chen}{ibm}
\icmlauthor{Sy-Yen Kuo}{ntu}
\end{icmlauthorlist}

\icmlaffiliation{duke}{Department of Electrical and Computer Engineering, Duke University, Durham, NC, USA}
\icmlaffiliation{ntu}{Department of Electrical Engineering, National Taiwan University, Taipei, Taiwan}
\icmlaffiliation{nycu}{Department of Electronics and Electrical Engineering,
National Yang Ming Chiao Tung University Hsinchu City, Taiwan}
\icmlaffiliation{ibm}{IBM Research, New York, USA}

\icmlcorrespondingauthor{Ming-Yu Chung}{ming-yu.chung@duke.edu}

\icmlkeywords{Model Reprogramming, Neural Tangent Kernel}

\vskip 0.3in
]
\printAffiliationsAndNotice{}  

\begin{abstract}
Model Reprogramming (MR) is a resource-efficient framework that adapts large pre-trained models to new tasks with minimal additional parameters and data, offering a promising solution to the challenges of training large models for diverse tasks. Despite its empirical success across various domains such as computer vision and time-series forecasting, the theoretical foundations of MR remain underexplored. In this paper, we present a comprehensive theoretical analysis of MR through the lens of the Neural Tangent Kernel (NTK) framework. We demonstrate that the success of MR is governed by the eigenvalue spectrum of the NTK matrix on the target dataset and establish the critical role of the source model's effectiveness in determining reprogramming outcomes. Our contributions include a novel theoretical framework for MR, insights into the relationship between source and target models, and extensive experiments validating our findings.
\end{abstract}

\section{Introduction}\label{sec: introduction}
In recent years, machine learning has achieved remarkable success in various domains, including computer vision (CV), medical AI, and large language models (LLMs). However, these advancements often rely on models with an enormous number of parameters, sometimes exceeding a billion or more. Ensuring the performance of specific tasks necessitates training such large models on vast datasets. Given this scale, developing a machine learning pipeline or algorithm becomes an expensive and time-intensive endeavor, as exemplified by ChatGPT-3 \citep{brown2020language}. When faced with numerous tasks, it becomes economically and temporally impractical to train multiple models.

\begin{figure}[t]
    \centering
    \centerline{\includegraphics[width=0.8\linewidth]{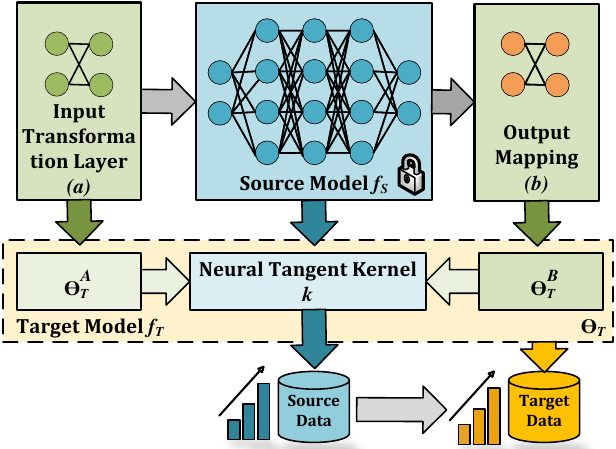}}
    \vspace{-10pt}
    \caption{Schematic illustration of Model Reprogramming (MR): The target model $f_T$ is composed with input transformation layer $a$, output mapping $b$ and source model $f_S$. For neural tangent kernel (NTK), $k(\cdot, \cdot)$ is the NTK induced by the source model $f_S$, while $\Theta_T^A$ and $\Theta_T^B$ are NTK induced by input transformation layer and output mapping respectively.   It is important to mention that the input transformation layer and output mapping are derived by optimizing Eq.~\ref{eq: MR}, while the source model is frozen during the optimization process. In this paper, we aim to utilize the NTK Theory to identify the relation between the source model and the target model.}
    \label{fig:MR}
\end{figure}
Given the substantial resources invested in developing large models for specific tasks, an intriguing question has emerged: can these costly large models be repurposed to tackle different tasks? This inquiry has led to the development of a technique known as Model Reprogramming (MR) \citep{chen2024model, yang2023english, yang2021voice2series,elsayed2018adversarial}, also referred to as Visual Prompt (VP) \citep{bahng2022exploring,chen2023understanding,tsao2023autovp}, specifically for computer vision tasks. MR is a machine learning framework that leverages an existing source model (a large model designed for a particular source task), a small amount of target data, and a minimal number of trainable parameters to create a model capable of addressing a new target task. According to \citep{chen2024model}, MR is often implemented by adding a trainable input transformation layer and an output mapping layer (either trainable or pre-specified) to a source model for reprogramming (see Figure~\ref{fig:MR} for the MR in the classification setting). Examples include reprogramming human acoustic models for time series classification \citep{yang2021voice2series}, reprogramming English language models for protein sequence learning \citep{melnyk2023reprogramming,vinod2023reprogramming}, and reprogramming LLMs for time-series forecasting \citep{jin2023time}, to name a few.

Despite the impressive empirical success of MR, most MR-like algorithms face two significant limitations: (1) the lack of holistic theoretical analysis, (2) insufficient investigation into the relationship between MR, target data, and source data. In this paper, we address these issues through the framework of the Neural Tangent Kernel (NTK) \citep{jacot2018neural, lee2019wide, huang2020dynamics, he2020bayesian}. We begin by describing the mechanics of MR within the NTK framework. Based on this NTK representation, we demonstrate that the success of MR can be characterized by the eigenvalues of the kernel matrix evaluated on the target dataset. Our analysis also explains why the success of MR often depends on the effectiveness of the source model, as has been empirically found in prior works such as \citep{tsao2023autovp, li2023exploring}. Furthermore, we support our theoretical findings with extensive experiments, showing that the fast algorithm is effective in real-world scenarios. The experimental results suggest that our theoretical prediction is valid.

Our contributions are summarized as follows. (1) We develop a theoretical framework for analyzing Model Reprogramming (MR) using the Neural Tangent Kernel (NTK). Specifically, our approach leverages the eigenvalue spectrum of the NTK matrix to characterize the success of MR. (2) We provide an explanation for the observed phenomenon that the success of MR is closely tied to the effectiveness of the source model \citep{tsao2023autovp, li2023exploring}. In particular, we identify the relationship between the eigenvalue spectrum of the source model and that of the target model, highlighting this connection as a critical factor underlying the phenomenon. (3) We perform comprehensive experiments to validate our theoretical findings.

\section{Preliminaries and Related Works}\label{sec: preliminaries}
\paragraph{Model Reprogramming}
We formally define Model Reprogramming (MR) in the context of classification tasks as follows. Given a source dataset \(D_S = \{(x_i, y_i)\}_{i=1}^{N_S}\) sampled from the source distribution \((x_s, y_s) \sim \mathfrak{D}_S\), where \(x_s \in \mathcal{X}_S \subset \mathbb{R}^{d_S}\) and \(y_s \in \mathcal{Y}_S \subset \mathbb{R}^{c_S}\), and a source model \(f_S : \mathbb{R}^{d_S} \rightarrow \mathbb{R}^{c_S}\) trained on the source dataset \(D_S\), along with a target dataset \(D_T = \{(x_i, y_i)\}_{i=1}^{N_T}\) sampled from the target distribution \((x_t, y_t) \sim \mathfrak{D}_T\), where \(x_t \in \mathcal{X}_T \subset \mathbb{R}^{d_T}\) and \(y_t \in \mathcal{Y}_T \subset \mathbb{R}^{c_T}\), we aim to solve the following optimization problem:
\begin{equation}\label{eq: MR}
    a^*, b^* 
    = {\arg\min}_{a\in\mathcal{H}_A, b\in\mathcal{H}_B}
    \mathbb{E}_{(x, y)\sim D_T} \|b\circ f_S \circ a (x) - y\|_2^2.
\end{equation}

Here, $\mathcal{H}_A \subset \{a: \mathbb{R}^{d_T} \rightarrow \mathbb{R}^{d_S}\}$ and $\mathcal{H}_B \subset \{b: \mathbb{R}^{c_S} \rightarrow \mathbb{R}^{c_T}\}$ are the hypothesis classes (neural networks) for the input transformation layer and the output mapping, respectively. The optimal solutions $a^*$ and $b^*$ are selected as the input transformation layer and the output mapping, respectively. MR generates a target model $f_T: \mathbb{R}^{d_T} \rightarrow \mathbb{R}^{c_T}$ defined as $b^* \circ f_S \circ a^*$ to solve the target problem. The proposed target model $f_T$ is expected to achieve high accuracy on the target distribution $\mathfrak{D}_T$. The schematic diagram of MR is demonstrated in Figure~\ref{fig:MR}.

\begin{figure}[t]
    \centering
    \centerline{\includegraphics[width=0.8\linewidth]{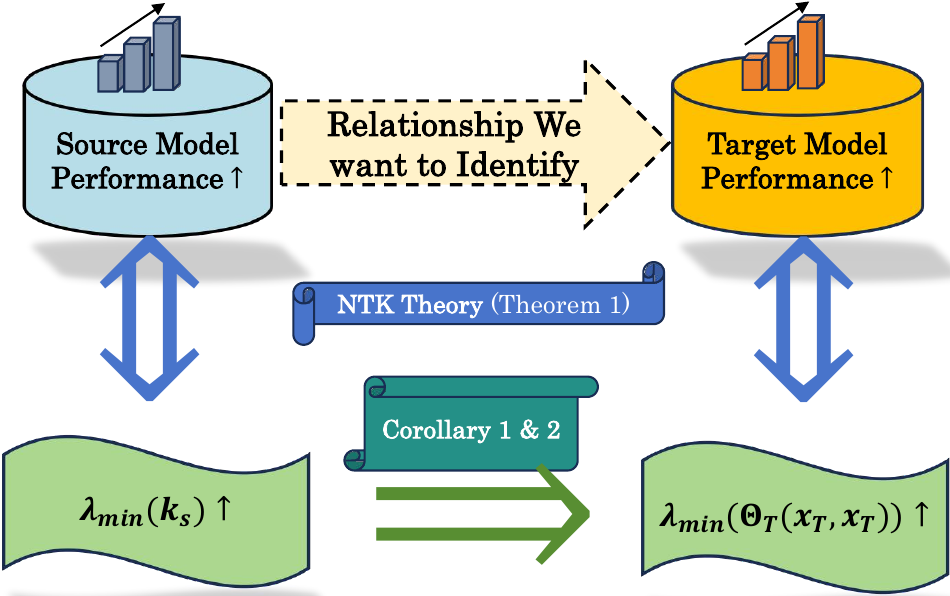}}
    \vspace{-10pt}
    \caption{This graph demonstrate our thinking for Model Reprogramming (MR). In this paper, we aims to identify the relationship between performance of source model and performance of target model. Our strategy is that, by NTK Theory, we first relate the performance of model with the minimum eigenvalue of its kernel matrix. Then, we bridge the relationship between the source model and target model by proving the proportional relation between minimum eigenvalues corresponding to the source model and the target model, under some additional assumptions.}
    \label{fig: MR_2}
\end{figure}

Recently, \citet{chen2024model} provided a comprehensive survey of MR. MR has achieved significant success in different data domains. For continuous data domains, \citet{yang2021voice2series} reprogrammed a speech model for time-series classification problems. \citet{tsai2020transfer} improved the settings of adversarial reprogramming for black-box models. For discrete data domains, \citet{vinod2023reprogramming} investigated representation reprogramming through dictionary learning, while \citet{hambardzumyan2021warp} studied adversarial reprogramming in word-level contexts.

Regarding the theory of MR, \citet{yang2021voice2series} roughly bounded the risk of the target model by the summation of source risk and the Wasserstein metric, which measures the distance between the target distribution and the source distribution. \citet{englert2022adversarial} proved that the accuracy can be arbitrarily high when adversarially reprogramming some 2-layer ReLU neural networks. Unfortunately, these works could not fully characterize the mechanism of MR due to rigorous constraints and overly rough approximations.

\paragraph{Reproducing Kernel Hilbert Space}
One challenge in studying the mechanism of MR lies in the unclear relationship between the source dataset \(D_S\) and the source model \(f_S\). This is due to the training process being typically intractable and dependent on numerous hyperparameters~\citep{chen2024model, yang2023english, tsao2023autovp, yang2021voice2series}. To alleviate this issue, in our study, we assume that the hypothesis class of the source model \(\mathcal{H}_S\) is a \textit{Reproducing Kernel Hilbert Space}~(RKHS)~\citep{aronszajn1950theory, berlinet2011reproducing, ghojogh2021reproducing}, which is defined as follows::
\begin{definition}[Kernel]\label{def:kernel}
A real value binary operation $k: \mathcal{X} \times \mathcal{X} \rightarrow\mathbb{R}$ is a kernel if $k(\cdot, \cdot)$ satisfies
\begin{itemize}
    \item (Symmetric): $k(x, x') = k(x', x)$ for all $x, x' \in\mathcal{X}$,
    \item (Positive semi-definite): For any sequence $X = \{x_1, x_2, \dots, x_n\}$ where $x_i \in \mathcal{X}$. The kernel matrix $K$ noted as $k(X, X)$ is positive semi-definite where $K_{i, j} := k(x_i, x_j)$.
\end{itemize}
\end{definition}

\begin{definition}[Reproducing Kernel Hilbert Space]\label{def:rkhs}
Given a kernel $k:\mathcal{X}\times\mathcal{X}\rightarrow\mathbb{R}$, the hypothesis class $\mathcal{H}_{k}=\{f:\mathcal{X}\rightarrow\mathbb{R}\}$ is called a RKHS, if $\mathcal{H}_k$ satisfies
\begin{itemize}
    \item  $k(\cdot, x) \in \mathcal{H}_{k}$,~$\forall x\in\mathcal{X}$,
    \item $\mathcal{H}_k$ is a Hilbert space built on the inner product $\langle\cdot, \cdot\rangle_{\mathcal{H}_k}$,
    \item (Reproducing property): $\forall x\in\mathcal{X}$ and $\forall f\in\mathcal{H}_{k}$, $f(x) = \langle f, k(\cdot, x)\rangle_{\mathcal{H}_k}$. 
\end{itemize}
\end{definition}

Under this assumption, with the help of Representer Theorem \citep{kimeldorf1971some,ghojogh2021reproducing} induced by the reproducing property, the model trained on source dataset $f_S$ possesses closed-form solution. Suppose the training process of the source model is defined as following optimization problem
\begin{equation}
    f_S = {\arg\min}_{f\in\mathcal{H}_S} \frac{1}{N_S}\sum_{(x_s, y_s)\in D_S} \|f(x_s) - y_s\|_2^2 + \sigma\|f\|^2_{\mathcal{H}_S},
\end{equation}
where $\sigma>0$ is regularization parameter and $\|\cdot\|_{\mathcal{H}_S}$ is the norm corresponding to the inner product $\langle\cdot, \cdot\rangle_{\mathcal{H}_S}$ defined in Definition \ref{def:rkhs}, 
the closed-form expression of $f_S$ is depicted as 
\begin{equation}\label{sol:c-closed-form}
    f_{S}(\cdot)^T = k(\cdot, X_{S}) [k(X_{S}, X_{S}) + \sigma\cdot N_{S}\cdot I]^{-1} Y_{S}, 
\end{equation}
where $X_S = [x_s^1, x_s^2, \dots, x_s^{N_S}]^T$ is a $N_S \times d_S$ matrix, $Y_S = [y_s^1, y_s^2, \dots, y_s^{N_S}]^T$ is a $N_S \times c_S$ matrix, $N_S$ is the sample size of the source dataset, $I$ is a $N_S \times N_S$ identity matrix, and $k(\cdot, \cdot)$ is the kernel defined in Definition~\ref{def:kernel}. An important question is how to construct the kernel $k(\cdot, \cdot)$ such that treating $\mathcal{H}_S$ as a Reproducing Kernel Hilbert Space (RKHS) is a valid assumption. In this paper, we consider the kernel $k(\cdot, \cdot)$ to be induced by the \textit{Neural Tangent Kernel} (NTK), which will be introduced in the next paragraph.

\paragraph{Neural Tangent Kernel} \textit{Neural Tangent Kernel}~(NTK) is first introduced by \citet{jacot2018neural} and  defined below.
\begin{definition}[Neural Tangent Kernel]\label{def: NTK}
Given a hypothesis class $\mathcal{H} = \{f_{\theta}:\mathbb{R}^d \rightarrow \mathbb{R}^c | \theta \in \mathbb{R}^p\}$ and a model $f=[f^1, f^2, \dots, f^c]\in\mathcal{H}$, the corresponding NTK, noted as $\hat\Theta$, is defined as 
\begin{equation}
    \hat\Theta^{i, j} (x, x') = \langle\nabla_{\theta}f^i(x), \nabla_{\theta}f^j(x')\rangle_{\mathbb{R}^p}.
\end{equation}
\end{definition}

In the context of using gradient descent as the optimization algorithm, \citet{jacot2018neural} demonstrated that as the width of a neural network approaches infinity, the NTK $\hat\Theta(x, x')$ converges to a kernel defined in Definition~\ref{def:kernel}. Specifically,
\begin{equation}\label{eq: NTK->kernel}
    \hat\Theta(x, x') \rightarrow \Theta(x, x')\cdot I_c,
\end{equation}
where $\Theta(x, x')\in\mathbb{R}$ and $I_c$ is a $c \times c$ identity matrix. Additionally, \citet{jacot2018neural} proved the equivalence between infinite-width neural networks and kernel regression in RKHS. In other words, if the neural network's width is sufficiently large, the model (infinite-width neural network) trained on the training set $(X, Y)$ can be approximated as
\begin{equation}
    \Theta(\cdot, X)\Theta^{-1}(X, X) Y\label{eq: NTK_0},
\end{equation}
where $X$ is an $N \times d$ matrix, $Y$ is an $N \times c$ matrix, and $N$ is the number of training samples. To ensure the invertibility of $\Theta(X, X)$ and to address the convergence issues caused by the rapid decay of the NTK eigenvalue spectrum~\citep{ronen2019convergence}, Eq.~\ref{eq: NTK_0} is typically modified as 
\begin{equation}
    \Theta(\cdot, X)[\Theta(X, X) + \sigma I]^{-1} Y.\label{eq: NTK_1}
\end{equation}
This modification is equivalent to the closed-form solution mentioned in RKHS (Eq.~\ref{sol:c-closed-form}).

Since the introduction of NTK, numerous papers have utilized NTK theory. \citet{lee2019wide} demonstrated that wide neural networks evolve as linear models and that NTK remains constant during the training process. In the overparameterized regime, \citet{du2019gradient} studied optimization convergence, while \citet{arora2019fine} investigated generalization. \citet{he2020bayesian} explored the relationship between deep ensembles and Gaussian processes. \citet{huang2020dynamics} proposed the \textit{Neural Tangent Hierarchy} (NTH) to study the dynamics of deep neural networks. \citet{nguyen2020dataset} used NTK to develop a novel dataset condensation technique, and \citet{chung2024rethinking} examined backdoor attacks on dataset distillation under NTK scenarios. Recently, \citet{chen2024analyzing} introduced the \textit{Loss Path Kernel} (LPK) to derive a tight generalization bound.

\section{Theoretical Framework}\label{sec: theory}
In this paper, we aim to analyze the success of MR. The success of MR is represented by the risk of the target model $f_T$ evaluated on the target distribution $\mathfrak{D}_T$:
\begin{equation}\label{eq: MR Risk}
    \mathbb{E}_{(x_t, y_t)\sim\mathfrak{D}_T} \|f_T(x_t) - y_t\|_2^2.
\end{equation}
Here, $f_T = b^* \circ f_S \circ a^*$, where $f_S$ is the source model, and $a^*, b^*$ are derived by solving Eq.~\ref{eq: MR}. In the classical framework, risk can be divided into two parts: empirical risk and generalization gap. This is expressed as follows:
\begin{small}
\begin{align}\label{eq: framework}
&\mathbb{E}_{(x_t, y_t)\sim\mathfrak{D}_T} \|f_T(x_t) - y_t\|_2^2 \nonumber\\
&=
\underbrace{\mathbb{E}_{(x_t, y_t)\sim D_T} \|f_T(x_t) - y_t\|_2^2}_{\text{Empirical Risk}} \nonumber\\
&+
\underbrace{[\mathbb{E}_{(x_t, y_t)\sim\mathfrak{D}_T} \|f_T(x_t) - y_t\|_2^2
-
\mathbb{E}_{(x_t, y_t)\sim D_T} \|f_T(x_t) - y_t\|_2^2]}_{\text{Generalization Gap}}.
\end{align}
\end{small}
It is important to note that $\mathfrak{D}_T$ is the target distribution and $D_T$ is the target dataset sampled from $\mathfrak{D}_T$. To clearly relate $f_T$ and $D_T$, we assume that $f_T$ is a neural network of sufficient width. Therefore, we can express $f_T$ in NTK form as follows:
\begin{equation}\label{eq: f_T NTK}
    f_T (\cdot) \approx \Theta_T (\cdot, X_T) [\Theta_T (X_T, X_T) + \sigma I]^{-1} Y_T, 
\end{equation}
where $\Theta_T$ is the corresponding NTK of $f_T$, $(X_T, Y_T)$ is the target dataset, and $\sigma > 0$ is a given regularization parameter. Under this analytical framework (Eq.~(\ref{eq: framework})) and NTK assumption (Eq.~\ref{eq: f_T NTK}), we can derive the upper bound of the empirical risk and the generalization gap to represent the success of MR, which will be discussed in the remainder of this section.

\subsection{Empirical Risk \& Generalization Gap of MR}
First, we analyze the empirical risk. Given a target dataset $D_T = \{(x_i, y_i)\}_{i=1}^{N_T}$ and a regularization parameter $\sigma > 0$, and considering the NTK expression of the target model $f_T$, the empirical risk $\mathcal{L}_{\text{ER}}$ is given by:
\begin{align}\label{eq: def of ER}
    &\mathcal{L}_{\text{ER}} 
    = 
    \mathbb{E}_{(x_t, y_t)\sim D_T} \|f_T (x_t) - y_t\|_2^2 \nonumber\\
    &=
    \frac{1}{N_T} \|\{I - \Theta_T (X_T, X_T) [\Theta_T (X_T, X_T) + \sigma I]^{-1}\} Y_T \|_2^2.
\end{align}
Through further analysis, the empirical risk $\mathcal{L}_{\text{ER}}$ can be bounded by the theorem below.
\begin{theorem}[Bound of Empirical Risk]\label{thm: empirical risk}
Let the eigenvalues of the kernel matrix $\Theta_T(X_T, X_T)$ be denoted as $\{ \lambda_i \}_{i=1}^{N_T}$, where $\lambda_i \geq \lambda_j$ for all $i < j$. The empirical risk $\mathcal{L}_{\text{ER}}$ can be bounded by:
\begin{equation}
\frac{1}{N_T}[1 - \frac{\lambda_{1}}{\sigma + \lambda_{1}}]\cdot\|Y_T\|_2^2
\leq 
\mathcal{L}_{\text{ER}} 
\leq 
\frac{1}{N_T}[1 - \frac{\lambda_{N_T}}{\sigma + \lambda_{N_T}}]\cdot\|Y_T\|_2^2.
\end{equation}
\end{theorem}

The proof of Theorem~\ref{thm: empirical risk} can be found in Appendix~\ref{sec: Theorem 1 and Its Proof}. This theorem indicates that the empirical risk $\mathcal{L}_{\text{ER}}$ is influenced by the eigenvalue spectrum of $\Theta_T(X_T, X_T)$. As the minimum eigenvalue $\lambda_{N_T}$ increases, $[1 - \frac{\lambda_{N_T}}{\sigma + \lambda_{N_T}}]$ approaches $0$, implying that the empirical risk $\mathcal{L}_{\text{ER}}$ decreases to zero. Therefore, to minimize the empirical risk $\mathcal{L}_{\text{ER}}$, it is essential to adjust the structure of $f_T$ such that the kernel matrix $\Theta_T(X_T, X_T)$ has a larger minimum eigenvalue.

On the other hand, for the generalization gap, the corresponding analysis  is provided in the Appendix~\ref{sec: Generalization Gap of MR}. We can prove that generalization gap is dominated by the Forbenius norm of $\Theta_T(X_T, X_T)$ and $\Gamma_{\mathfrak{D}_T}$ which is a large constant such that
\begin{equation}
    \|(x, y) - (x', y')\|_2 \leq\Gamma_{\mathfrak{D}_T},~\forall (x, y), (x', y')\sim\mathfrak{D}_T.
\end{equation}
Notice that $\Gamma_{\mathfrak{D}_T}$ is usually be supposed as a very large number, so that 
\begin{equation}
    \Gamma_{\mathfrak{D}_T} >> \|\Theta_T (X_T, X_T)\|_F.
\end{equation} 
Based on our analysis in Appendix~\ref{sec: Generalization Gap of MR}, this implies the generalization gap could be insensitive when we slightly vary the NTK of the target model. Thus, in this paper, we will focus on using eigenvalue spectrum to characterize the performance of Model Reprogramming (MR).

\paragraph{Summary and Remark} 
As discussed in previous sections, the eigenvalue spectrum of the kernel matrix $\Theta_T (X_T, X_T)$ plays a crucial role in determining the success of MR, which is supported by previous work~\cite{du2019gradient}.
One might intuitively assume that $\Theta_T (X_T, X_T)$ is solely dependent on the neural network structure and the target distribution $\mathfrak{D}_T$. However, $\Theta_T (X_T, X_T)$ is also influenced by the source distribution $\mathfrak{D}_S$. This is because the target model $f_T = b^* \circ f_S \circ a^*$, where $f_S$ is the model trained on the source dataset $D_S$. Therefore, the behavior of $\Theta_T (X_T, X_T)$ is an interplay between the model structure, the target distribution $\mathfrak{D}_T$, and the source distribution $\mathfrak{D}_S$. To gain a deeper understanding of the MR mechanism, it is essential to examine the impact of $\mathfrak{D}_S$ on $\Theta_T (X_T, X_T)$, which will be discussed in the next section.

\section{Relation between NTK, Target Distribution, and Source Distribution}
\label{sec: relation target and source}
In this section, we analyze the relationship between $\Theta_T (X_T, X_T)$, the target distribution $\mathfrak{D}_T$, and the source distribution $\mathfrak{D}_S$. Our analysis framework is visualized in the Fig.~\ref{fig: MR_2}. Recall that the target model $f_T$ is constructed based on the structure $b \circ f_S \circ a$, where $f_S$ is a pre-existing (untrainable) source model. To simplify the analysis, we suppose 
Assumption~\ref{assump: ntk} holds.
\begin{assumption}\label{assump: ntk}
In order to clearly express the relation between $\Theta_T (X_T, X_T)$, target distribution $\mathfrak{D}_T$ and source distribution $\mathfrak{D}_S$, we consider the following assumptions:
\begin{enumerate}
        \item The source model $f_S$ can be expressed by some kernel model. Namely,
        \begin{equation}
            [f_S (\cdot)]^T = k(\cdot, X_S)[K_S + \sigma_S I]^{-1} Y_S,
        \end{equation}
        where the kernel matrix $K_S = k(X_S, X_S)$, $\sigma_S > 0$ is regularization parameter and the kernel $k(x, x') = \langle \Phi(x), \Phi(x')\rangle$ is induced by NTK.
        \item $b\in\mathcal{H}_B$ is a $c_T \times c_S$ linear matrix and $b = [b^1, b^2, \dots, b^{c_T}]^T$, $b^i \in\mathbb{R}^{c_S}$. Specifically, the hypothesis class $\mathcal{H}_B = \{b~|~\mathbb{R}^{c_T \times c_S}\}$.
\end{enumerate}
\end{assumption}

From Theorem~\ref{thm: empirical risk}, we know that the eigenvalue spectrum of $\Theta_T(X_T, X_T)$ characterizes the performance of the target model $f_T$ when evaluated on the target distribution $\mathfrak{D}_T$. Thus, to better understand the mechanism of MR, it is essential to analyze MR in relation to the eigenvalue spectrum of $\Theta_T(X_T, X_T)$. However, in most cases, deriving the explicit formulation of $\Theta_T(X_T, X_T)$ is highly challenging. On the other hand, representing $\hat\Theta_T(X_T, X_T)$ is significantly easier than representing $\Theta_T(X_T, X_T)$. To address this issue, we prove Proposition~\ref{prop: spectrum}, which shows that $\Theta(X_T, X_T)$ and $\hat\Theta(X_T, X_T)$ share the same eigenvalue spectrum. Thus, in our subsequent analysis, we can focus on $\hat\Theta_T(X_T, X_T)$ when the explicit form of $\Theta(X_T, X_T)$ is difficult to derive.
\begin{proposition}\label{prop: spectrum}
    Assume that the width of the target model $f_T: \mathbb{R}^{d_T}\rightarrow\mathbb{R}^{c_T}$ is sufficient large and hence $\hat\Theta_T(x, x') \rightarrow \Theta_T(x, x') I_{c_T}$, then the eigenvalue spectrum of $\hat\Theta_T (X_T, X_T)$ is equivalent to the eigenvalue spectrum of $\Theta_T(X_T, X_T)$. To be more specific, denote $\{\lambda_i\}_{i=1\sim N_T}$ as the eigenvalue spectrum of $\Theta_T(X_T, X_T)$, the eigenvalue spectrum of $\hat\Theta_T (X_T, X_T)$ will be $\{\lambda_i^j\}_{i=1\sim N_T \atop j=1\sim c_T}$ where $\lambda_i^j = \lambda_i$ for all $i$ and $j$.
\end{proposition}
Proposition~\ref{prop: spectrum} is derived from the properties of the tensor product~$\otimes$. We observe that $\hat\Theta(X_T, X_T)$ and $\Theta(X_T, X_T)$ are related through the tensor product $\otimes$~\citep{jacot2018neural}. Consequently, we can establish a connection between the eigenvalue spectra of $\hat\Theta(X_T, X_T)$ and $\Theta(X_T, X_T)$. The proof of Proposition~\ref{prop: spectrum} can be found in Appendix~\ref{sec: Proposition 1 and Its Proof}.

Finally, our analysis is motivated by the following observation. Given a target model $f_T(\cdot) = b \circ f_S \circ a(\cdot)$, where $\theta_A$ parameterizes the hypothesis class $\mathcal{H}_A$ (Input Transformation Layer) and $\theta_B$ parameterizes the hypothesis class $\mathcal{H}_B$ (Output Mapping), we find that the NTK of the target model $f_T(\cdot)$ can be expressed as the sum of the NTK induced by the input transformation layer and the NTK induced by the output mapping. That is,
\begin{small}
\begin{align}
    &\hat\Theta_T(x, x')
    =
    \nabla_{\theta_T} f_T(x) [\nabla_{\theta_T} f_T(x')]^T \\
    &=
    \nabla_{\theta_A} f_T(x) [\nabla_{\theta_A} f_T(x')]^T 
    +
    \nabla_{\theta_B} f_T(x) [\nabla_{\theta_B} f_T(x')]^T \\
    &=
    \hat\Theta_T^A (x,x') + \hat\Theta_T^B (x,x'),\label{eq: NTKdecom}
\end{align}
\end{small}
where $\theta_T = (\theta_A, \theta_B)$ is denoted as all trainable parameters of the target model $f_T$, $\hat\Theta_T^A (X_T, X_T)$ is the NTK induced by input transformation layer, and $\hat\Theta_T^B (X_T, X_T)$ is the NTK induced by output mapping. This decomposition inspires us to divide our analysis structure into two stages: (1.) Analyze the eigenvalue spectrum of $\hat\Theta_T^A (X_T, X_T)$
and $\hat\Theta_T^B (X_T, X_T)$ respectively. (2.) Utilize the information derived in previous two stages to infer the properties of the eigenvalue spectrum of $\Theta_T(X_T, X_T)$.
The details are presented in the subsequent subsections.

\subsection{NTK Induced by Input Transformation Layer}\label{sec: NTK_A}

First, we consider the Jacobian matrix corresponding to $\theta_A$
\begin{small}
\begin{align}
    &\nabla_{\theta_A} f_T(x) 
    = 
    \nabla_{\theta_A} b f_S(a(x))
    \nonumber\\
    &=
    b Y_S^T [K_S + \sigma_S I]^{-1} \Phi(X_S)^T \nabla_{a} \Phi(a(x)) \nabla_{\theta_A} a(x).\label{eq: hessian_A}
\end{align}
\end{small}

Then, following the definition of NTK~(Definition~\ref{def: NTK}) and Eq.~\ref{eq: hessian_A}, $\hat\Theta_T^A (x, x')$ can be formulated as
\begin{align}
    \hat\Theta_T^A (x, x')
    =
    \nabla_{\theta_A} f_T(x) \cdot [\nabla_{\theta_A} f_T(x')]^T.
    \label{eq: ntk_mr}
\end{align}
With Eq.~\ref{eq: ntk_mr}, we can derive the following theorem to characterize the eigenvlaue spectrum of $\hat\Theta_T^A(X_T, X_T)$.

\begin{theorem}\label{thm: spectrum}
Suppose Assumption~\ref{assump: ntk} holds, the eigenvalue spectrum of the kernel matrix $\hat\Theta_T^A(X_T, X_T)$ can be bounded as follows.
\begin{footnotesize}
\begin{align}
    &\lambda_i (\hat\Theta_T^A(X_T, X_T))
    \nonumber\\ 
    &\leq
    \lambda_{\text{max}}[\Theta_S^b]
    \cdot
    \sup_{(x_t, y_t) \in D_T}\lambda_{\text{max}}[\hat\Theta_S^A(x_t, x_t)]\cdot 
    \lambda_{\text{max}}[\hat\Theta_A(X_T, X_T)]
\end{align}
\end{footnotesize}
and
\begin{footnotesize}
\begin{align}
    &\lambda_i (\hat\Theta_T^A(X_T, X_T))
    \nonumber\\
    &\geq
    \lambda_{\text{min}}[\Theta_S^b] 
    \cdot\inf_{(x_t, y_t) \in D_T}\lambda_{\text{min}}[\hat\Theta_S^A(x_t, x_t)]\cdot \lambda_{\text{min}}[\hat\Theta_A(X_T, X_T)]
\end{align}
\end{footnotesize}
where $\lambda_i(\cdot)$ is the operator to output the $i$-th large eigenvalue,
$\hat\Theta^A(x, x') = \nabla_{\theta_A} a(x) [\nabla_{\theta_A} a(x')]^T$, $\hat\Theta^A_S(x, x') = \nabla_a \Phi(a(x)) [\nabla_a \Phi(a(x'))]^T$, $\Theta_S^b = b Y_S^T [K_S + \sigma_S I]^{-1} K_S [K_S + \sigma_S I]^{-1} Y_S b^T$.
\end{theorem}
The proofs of Theorem~\ref{thm: spectrum} can be found in Appendix~\ref{sec: Theorem 3 and Its Proof}.  
In Theorem~\ref{thm: spectrum}, if we fix the structure of the input transformation layer and vary the source model, we observe that the eigenvalue spectrum of $\hat\Theta_T^A(X_T, X_T)$ is governed by $\hat\Theta_S^A(x_t, x_t)$, which also implicitly indicates the relationship between $\mathfrak{D}_T$ and $\mathfrak{D}_S$. Furthermore, we identify a sufficient condition (Assumption~\ref{assump: cor}) under which the minimum eigenvalue of $\hat\Theta_T^A(X_T, X_T)$ is proportional to the minimum eigenvalue of $K_S$, as stated in Corollary~\ref{cor: spectrum}. Assumption~\ref{assump: cor} and Corollary~\ref{cor: spectrum} are presented below.

\begin{assumption}\label{assump: cor}
Given a source dataset $D_S$ and a target dataset $D_T$, there exists $c_A >0$ such that
\begin{equation}
    \lambda_{\text{min}}[\hat\Theta^A_S(x_t, x_t)] \geq c_A 
    \cdot
    (\lambda_{\text{max}}[K_S]+\sigma_S)~,\forall (x_t, y_t)\in D_T.
\end{equation}
\end{assumption}

\begin{corollary}\label{cor: spectrum}
Suppose Assumption~\ref{assump: ntk} and Assumption~\ref{assump: cor} hold, then we have
\begin{align}
    &\lambda_i (\hat\Theta_T^A(X_T, X_T))\nonumber\\ 
    &\geq
    \lambda_{\text{min}}(b Y_S^T Y_S b^T)
    \cdot
    [\frac{    
    c_A\cdot\lambda_{\text{min}}[K_S]}{\lambda_{\text{min}}[K_S] + \sigma}]
    \cdot
    \lambda_{\text{min}}[\hat\Theta_A(X_T, X_T)].
\end{align}
\end{corollary}
The proofs of Corollary~\ref{cor: spectrum} can be found in Appendix~\ref{sec: Corollary 1 and Its Proof}. Notably, if the output mapping is not trainable, Corollary~\ref{cor: spectrum} explains why the success of MR often depends on the success of the source model $f_S$, as observed in \citep{tsao2023autovp, li2023exploring}. Under this setting (where $\theta_B$ is fixed), we have $\hat\Theta_T(X_T, X_T) = \hat\Theta_T^A(X_T, X_T)$. Therefore, according to Corollary~\ref{cor: spectrum}, a larger $\lambda_{\text{min}}[K_S]$ leads to a larger $\lambda_{\text{min}}[\Theta_T(X_T, X_T)]$, which implies a lower empirical risk, as demonstrated in Theorem~\ref{thm: empirical risk}. At the same time, increasing $\lambda_{\text{min}}[K_S]$ reduces the empirical risk of the source model (as shown in Theorem~\ref{thm: empirical risk} by substituting $D_T$ with $D_S$ and $\Theta_T$ with the NTK induced by the source model). In summary, when $\lambda_{\text{min}}[K_S]$ is sufficiently large, both the empirical risk of the target model $f_T$ and the empirical risk of the source model $f_S$ converge to zero.

\subsection{NTK Induced by Output Mapping}\label{sec: NTK_B}
For output mapping, the NTK induced by $\theta_B$ would be
\begin{align}
    \hat\Theta_T^B (x, x') = [f_S(a(x))^T f_S (a(x'))] \otimes I_{c_T},
\end{align}
where $I_{c_T}$ is $\mathbb{R}^{c_T} \times\mathbb{R}^{c_T}$ identity matrix, and hence
\begin{align}
    \Theta_T^B(x, x') 
    &=  f_S(a(x))^T f_S (a(x'))\label{eq: hessian_B}.
\end{align}
Using Eq.~\ref{eq: hessian_B}, we can characterize the eigenvalue spectrum of $\Theta_T^B(X_T, X_T)$ as stated in the theorem below.

\begin{theorem}\label{thm: spectrum_B}
Suppose Assumption~\ref{assump: ntk} holds, the eigenvalue spectrum of the kernel matrix $\Theta_T^B(X_T, X_T)$ can be bounded as follows.
\begin{align}
    &\lambda_{i}(\Theta_B(X_T, X_T))
    \leq
    \lambda_{\text{max}}[k(a(X_T), X_S)k(X_S, a(X_T))]\nonumber\\
    &\cdot
    \lambda_{\text{max}}[[K_S + \sigma_S I]^{-2}]
    \cdot
    \lambda_{\text{max}}[Y_S Y_S^T]
\end{align}
and
\begin{align}
    &\lambda_{i}(\Theta_B(X_T, X_T))
    \geq
    \lambda_{\text{min}}[k(a(X_T), X_S)k(X_S, a(X_T))]\nonumber\\
    &\cdot
    \lambda_{\text{min}}[[K_S + \sigma_S I]^{-2}]
    \cdot
    \lambda_{\text{min}}[Y_S Y_S^T],
\end{align}
where $\lambda_i(\cdot)$ is the operator to output the $i$-th large eigenvalue.
\end{theorem}

The proofs of Theorem~\ref{thm: spectrum_B} can be found in Appendix~\ref{sec: Theorem 4 and Its Proof}. If we fix the input transformation layer and vary only the structure of the source model, Theorem~\ref{thm: spectrum_B} suggests that the eigenvalue spectrum of $\Theta_T^B(X_T, X_T)$ is determined by the eigenvalue spectrum of $K_S + \sigma_S I$ and $k(a(X_T), X_S)k(X_S, a(X_T))$, which characterize the relationship between $\mathfrak{D}_T$ and $\mathfrak{D}_S$. Further analysis reveals a sufficient condition (Assumption~\ref{assump: cor_2}) under which the eigenvalue spectrum of $\Theta_T^B(X_T, X_T)$ is proportional to the eigenvalue spectrum of $K_S$, as stated in Corollary~\ref{cor: spectrum_B}. Assumption~\ref{assump: cor_2} and Corollary~\ref{cor: spectrum_B} are presented below.

\begin{assumption}\label{assump: cor_2}
Given a source dataset $D_S$ and a target and dataset $D_T$, there exists $c_B > 0$ such that
\begin{equation}
    \lambda_{\text{min}}[k(a(X_T), X_S)k(X_S, a(X_T))] 
    \geq
    c_B \cdot
    (\lambda_{\text{max}}[K_S])^2.
\end{equation}
\end{assumption}

\begin{corollary}\label{cor: spectrum_B}
Suppose Assumption~\ref{assump: ntk}  and Assumption~\ref{assump: cor_2} hold, then we have
\begin{scriptsize}
\begin{align}
    \lambda_i (\Theta_T^B(X_T, X_T))
    \geq
    c_B\cdot[\frac
    {\lambda_{\text{min}}[K_S]}{\lambda_{\text{min}}[K_S] + \sigma_S I}]^2
    \cdot
    \lambda_{\text{min}}[Y_S Y_S^T].
\end{align}
\end{scriptsize}
\end{corollary}
The proof of Corollary~\ref{cor: spectrum_B} can be found in Appendix~\ref{sec: Corollary 2 and Its Proof}. Similar to the previous section, we can derive analogous discussions and results here. Specifically, if the input transformation layer is not trainable ($\theta_A$ is fixed), Corollary~\ref{cor: spectrum_B} explains why the success of MR often relies on the success of the source model $f_S$. Under this condition, we have $\Theta_T(X_T, X_T) = \Theta_T^B(X_T, X_T)$. Therefore, Corollary~\ref{cor: spectrum_B} implies that a larger $\lambda_{\text{min}}[K_S]$ leads to a larger $\lambda_{\text{min}}[\Theta_T(X_T, X_T)]$.
Thus, when $\lambda_{\text{min}}[\Theta_T(X_T, X_T)]$ is sufficiently large, both the empirical risk of the target model $f_T$ and the empirical risk of the source model $f_S$ converge to zero.

\subsection{NTK of the Target Model}{\label{sec: NTK_T}}
With the analysis in Sections~\ref{sec: NTK_A} and~\ref{sec: NTK_B}, we can characterize the eigenvalue spectrum of $\Theta_T(X_T, X_T)$. First, we notice that
\begin{small}
\begin{equation}
    \lambda_{i}[\hat\Theta_T(X_T, X_T)]
    \geq
    \lambda_{\min}\hat\Theta_T^A(X_T, X_T)  
    + 
    \lambda_{\min}\hat\Theta_T^B(X_T, X_T).
\end{equation}
\end{small}
By Proposition~\ref{prop: spectrum}, we know that $\lambda_{\text{min}}\hat\Theta_T^B(X_T, X_T)=\lambda_{\text{min}}\Theta_T^B(X_T, X_T)$ and $\lambda_{\text{min}}\hat\Theta_T(X_T, X_T)=\lambda_{\text{min}}\Theta_T(X_T, X_T)$. Thus, we have
\begin{align}
&\lambda_{i}[\Theta_T(X_T, X_T)]
\geq\nonumber\\
&\lambda_{\min}\hat\Theta_T^A(X_T, X_T)  
+ 
\lambda_{\min}\Theta_T^B(X_T, X_T).
\label{eq: NTK_T_1}
\end{align}
With similar derivation, we have
$\lambda_{\text{max}}\hat\Theta_T^B(X_T, X_T)=\lambda_{\text{max}}\Theta_T^B(X_T, X_T)$. Thus, we have
\begin{align}
&\lambda_{i}[\Theta_T(X_T, X_T)]
\leq\nonumber\\
&\lambda_{\max}\hat\Theta_T^A(X_T, X_T)  
+ 
\lambda_{\max}\Theta_T^B(X_T, X_T).
\label{eq: NTK_T_2}
\end{align}

Eqs.~\ref{eq: NTK_T_1} and~\ref{eq: NTK_T_2} indicate that the eigenvalue spectrum of $\Theta_T(X_T, X_T)$ can be governed by the combination of Theorem~\ref{thm: spectrum} and Theorem~\ref{thm: spectrum_B}. Moreover, Eqs.~\ref{eq: NTK_T_1} and~\ref{eq: NTK_T_2} also implicitly suggest that if either Assumption~\ref{assump: cor} or Assumption~\ref{assump: cor_2} holds, then $\lambda_{\min}[\Theta_T(X_T, X_T)]$ is proportional to $\lambda_{\min}[K_S]$. This explains why the success of MR depends on the success of the source model~\citep{tsao2023autovp} under the most general setting (i.e., both $\theta_A$ and $\theta_B$ are trainable).

\paragraph{Discussion and Implication of Assumptions}
Although Corollary~\ref{cor: spectrum} and Corollary~\ref{cor: spectrum_B} yield desirable results, the question of whether and when Assumption~\ref{assump: cor} and Assumption~\ref{assump: cor_2} hold remains open. To address this, we provide further discussion below.

The rationale behind Assumption~\ref{assump: cor} is discussed as follows. By definition, we know that $\hat\Theta^A_S(x_t, x_t) = \nabla_a \Phi(a(x_t)) [\nabla_a \Phi(a(x_t))]^T$. Intuitively, $\nabla_a \Phi(a(x_t))$ can be considered a feature extractor that maps target data $x_t$ to source features through the reprogramming layer $a$. The value $\lambda_{\text{min}}[\hat\Theta^A_S(x_t, x_t)]$ measures the redundancy of the feature extractor $\nabla_a \Phi(a(\cdot))$ when evaluated on the target dataset $D_T$. On the other hand, $\lambda_{\max}[K_S] = \lambda_{\max} [\Phi(X_S)^T\Phi(X_S)]$ seems to measure the performance of the feature extractor $\Phi(\cdot)$ when evaluated on $\mathfrak{D}_S$. Combining these two interpretations, we can infer that Assumption~\ref{assump: cor} requires the structure of the input transformation layer $\mathcal{H}_A$ to be sufficiently effective, such that the ability of the feature extractor $\nabla_a \Phi(a(\cdot))$ on $\mathfrak{D}_T$ aligns with the ability of the feature extractor $\Phi$ on $\mathfrak{D}_S$.  

However, while we can propose a potential explanation for Assumption~\ref{assump: cor}, examining it remains a challenge. Since $\Phi$ is not uniquely defined, it is impossible to search through all possible $\Phi$ in general. Thus, Assumption~\ref{assump: cor} is not practically identifiable.  

The rationale behind Assumption~\ref{assump: cor_2} is relatively straightforward. The value $\lambda_{\min}[k(X_T, X_S)k(X_S, X_T)]$ can be interpreted as a metric to measure the similarity between $\mathfrak{D}_T$ and $\mathfrak{D}_S$ through the lens of the source model. Assumption~\ref{assump: cor_2} requires that the structure of the input transformation layer $\mathcal{H}_A$ is sufficiently effective, such that the similarity between $\mathfrak{D}_T$ and $\mathfrak{D}_S$ aligns with the ability of the feature extractor $\Phi$ on $\mathfrak{D}_S$. Compared to Assumption~\ref{assump: cor}, Assumption~\ref{assump: cor_2} is identifiable. We conduct corresponding experiments to justify Assumption~\ref{assump: cor_2} in the section~\ref{sec: numerical results}.  

Finally, Assumption~\ref{assump: cor} and Assumption~\ref{assump: cor_2} implicitly suggest a criterion for the input transformation layer to ensure the scalability of MR. Intuitively, Assumption~\ref{assump: cor_2} can be used to identify a good structure for the input transformation layer to ensure the scalability of MR. The experimental results seem to support our conjecture. Nevertheless, further studies are needed.

\section{Numerical Results}\label{sec: numerical results}
In this section, we perform series of experiments to support our theoretical results. We aim to show that the minimum eigenvalue of the source model kernel matrix control both the loss of source model and the loss of the target model (Corollary~\ref{cor: spectrum} and Corollary~\ref{cor: spectrum_B}), which indicates the phenomenon that the success of MR usually relies on the success of the source model. Our experimental settings and corresponding results are presented below.
\subsection{Experimental Setting}
\paragraph{Dataset} We utilized ImageNet-10 as the source dataset, while the target dataset was chosen from CIFAR-10 or SVHN.
\paragraph{Input Transformation Layer \& Output Mapping} Two structures of the input transformation layer were considered in our experiments:1) \textbf{Single Fully Connected (FC)} was chosen as input transformation layer. The input shape was $24\times24\times3$ while the output shape was $32\times32\times3$. 2) \textbf{Visual Prompt (VP)} was used as input transformation layer in our experiment. VP output a $32\times32\times3$ image with $24\times24\times3$ input image and $32\times32\times3 - 24\times24\times3$ trainable noises. For output mapping, we used a simple linear matrix.
\paragraph{Source Model Structure} Three structures of the output mapping was considered in our experiments: 1) CNN 2) VGG 3) ResNet. The detailed structures were listed in the Appendix~\ref{sec: exp source model}

Other experimental settings are listed in the Appendix~\ref{sec: exp details}.

\begin{figure}[t]
	\centering
		\centerline{\includegraphics[width=\linewidth]{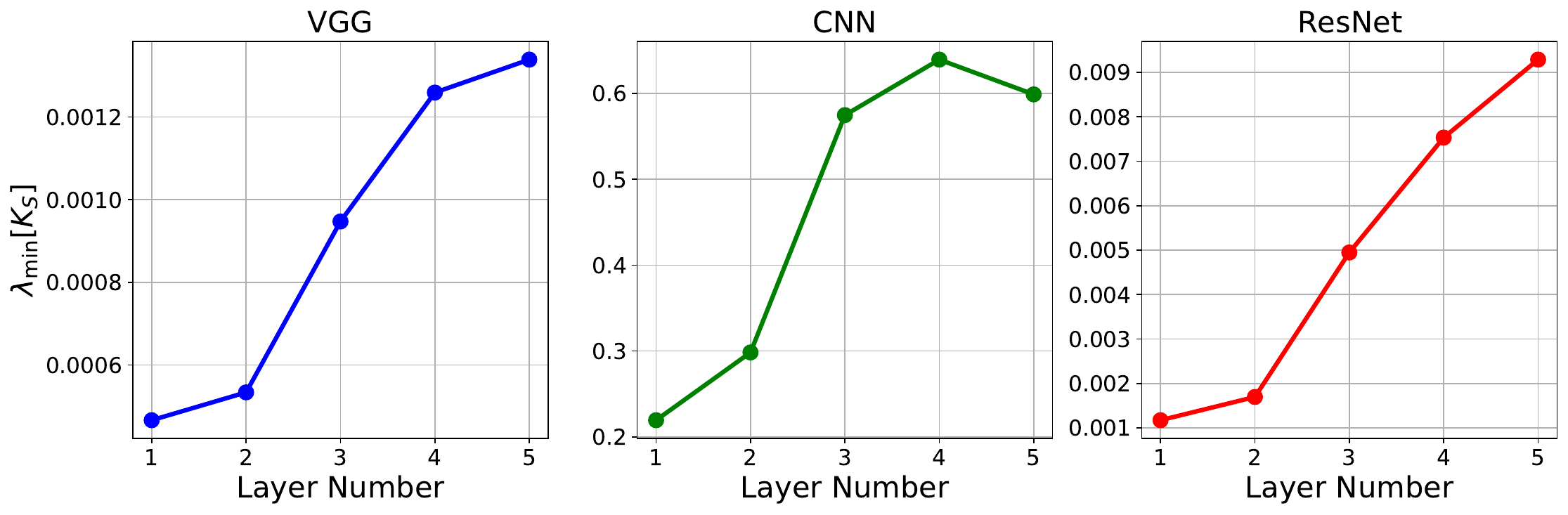}}
         \centerline{}
    \vspace{-20pt}
	\caption{$\lambda_{\min}[K_S]$ v.s. source model's depth for each source model} 
	\label{fig: NTKspectrum}
    \vspace{-10pt}
\end{figure}

\begin{table*}[t]
\caption{Source Data = ImageNet, Target Data = Cifar10. The cell values represent the $\text{Accuracy}_{\text{mean}}\pm\text{std} / \text{Loss}_{mean}\pm\text{std}$.}
\vspace{-10pt}
\centering
\resizebox{1\textwidth}{!}{
\begin{tabular}{c|c|c|c|c|c|c|c|c|c|c|c|c}
\toprule[1.5pt]
 & \multicolumn{6}{c|}{\textbf{FC}} &  \multicolumn{6}{c}{\textbf{VP}}\\
\midrule 
  & \multicolumn{2}{c|}{\textbf{VGG}} & \multicolumn{2}{c|}{\textbf{CNN}}& \multicolumn{2}{c|}{\textbf{RES}}& \multicolumn{2}{c|}{\textbf{VGG}} & \multicolumn{2}{c|}{\textbf{CNN}}& \multicolumn{2}{c}{\textbf{RES}}\\
 \midrule 
 Layer & Source & Target  & Source & Target  & Source & Target  & Source & Target  & Source & Target  & Source & Target \\
 \midrule 
   \midrule 
\cdashline{1-13}[2pt/2pt]
\rule{0pt}{10pt}
1 & 0.7215±0.0177/0.8431±0.0504 & 0.4335±0.0045/1.8302±0.0056 & 0.5406±0.0002/1.3186±0.0001 & 0.4050±0.0109/1.8196±0.0951 & 0.6640±0.0048/1.0289±0.0105 & 0.4340±0.0016/1.6908±0.0292 & 0.7215±0.0177/0.8431±0.0504 & 0.0978±0.0010/26.3892±0.5274 & 0.5406±0.0002/1.3186±0.0001 & 0.1516±0.0002/26.4519±0.0490 & 0.6640±0.0048/1.0289±0.0105 & 0.1003±0.0031/8.4982±0.9560 \\
\cdashline{1-13}[2pt/2pt]
\rule{0pt}{10pt}
2 & 0.7759±0.0017/0.7152±0.0054 & 0.4416±0.0009/1.7468±0.0175 & 0.6244±0.0014/1.1060±0.0007 & 0.3786±0.0185/2.0034±0.0902 & 0.6600±0.0062/1.0195±0.0055 & 0.4824±0.0022/1.4876±0.0157 & 0.7759±0.0017/0.7152±0.0054 & 0.1404±0.0018/18.3717±0.3750 & 0.6244±0.0014/1.1060±0.0007 & 0.1514±0.0007/26.6482±0.1783 & 0.6600±0.0062/1.0195±0.0055 & 0.1134±0.0070/6.6316±0.2121 \\
\cdashline{1-13}[2pt/2pt]
\rule{0pt}{10pt}
3 & 0.8014±0.0032/0.6448±0.0075 & 0.4675±0.0108/1.5688±0.0247 & 0.6835±0.0034/0.9741±0.0094 & 0.4287±0.0033/1.9241±0.0654 & 0.6798±0.0071/1.0373±0.0044 & 0.4717±0.0015/1.5112±0.0073 & 0.8014±0.0032/0.6448±0.0075 & 0.0693±0.0026/13.4800±0.6320 & 0.6835±0.0034/0.9741±0.0094 & 0.1271±0.0032/19.9919±0.5019 & 0.6798±0.0071/1.0373±0.0044 & 0.1573±0.0072/4.7034±0.4140 \\
\cdashline{1-13}[2pt/2pt]
\rule{0pt}{10pt}
4 & 0.8044±0.0025/0.6360±0.0016 & 0.4672±0.0065/1.5486±0.0338 & 0.7106±0.0002/0.8630±0.0045 & 0.4535±0.0049/1.7232±0.0067 & 0.6663±0.0049/1.0347±0.0146 & 0.4694±0.0038/1.5423±0.0132 & 0.8044±0.0025/0.6360±0.0016 & 0.0953±0.0161/9.7058±0.8565 & 0.7106±0.0002/0.8630±0.0045 & 0.0929±0.0066/15.2357±1.0594 & 0.6663±0.0049/1.0347±0.0146 & 0.1264±0.0058/4.6532±0.1567 \\
\cdashline{1-13}[2pt/2pt]
\rule{0pt}{10pt}
5 & 0.7923±0.0019/0.6858±0.0013 & 0.4381±0.0141/1.6201±0.0476 & 0.7343±0.0038/0.8023±0.0005 & 0.4622±0.0000/1.5954±0.0051 & 0.6572±0.0013/1.0551±0.0306 & 0.4602±0.0007/1.5967±0.0115 & 0.7923±0.0019/0.6858±0.0013 & 0.1011±0.0096/5.9333±0.3624 & 0.7343±0.0038/0.8023±0.0005 & 0.1121±0.0014/13.2805±0.5853 & 0.6572±0.0013/1.0551±0.0306 & 0.1615±0.0075/4.8415±0.1707 \\
\bottomrule[1.5pt]
\end{tabular}}
\label{tab: SITC_new_new}
\end{table*}

\begin{table*}[t]
\vspace{-10pt}
\caption{Source Data = ImageNet, Target Data = SVHN. The cell values represent the $\text{Accuracy}_{\text{mean}}\pm\text{std} / \text{Loss}_{mean}\pm\text{std}$.}
\vspace{-10pt}
\centering
\resizebox{1\textwidth}{!}{
\begin{tabular}{c|c|c|c|c|c|c|c|c|c|c|c|c}
\toprule[1.5pt]
 & \multicolumn{6}{c|}{\textbf{FC}} &  \multicolumn{6}{c}{\textbf{VP}}\\
\midrule 
  & \multicolumn{2}{c|}{\textbf{VGG}} & \multicolumn{2}{c|}{\textbf{CNN}}& \multicolumn{2}{c|}{\textbf{RES}}& \multicolumn{2}{c|}{\textbf{VGG}} & \multicolumn{2}{c|}{\textbf{CNN}}& \multicolumn{2}{c}{\textbf{RES}}\\
 \midrule 
 Layer & Source & Target  & Source & Target  & Source & Target  & Source & Target  & Source & Target  & Source & Target \\
 \midrule 
   \midrule 
\cdashline{1-13}[2pt/2pt]
\rule{0pt}{10pt}
1 & 0.7215±0.0177/0.8431±0.0504 & 0.5075±0.0011/2.0908±0.0063 & 0.5406±0.0002/1.3186±0.0001 & 0.5206±0.0094/1.7121±0.0656 & 0.6640±0.0048/1.0289±0.0105 & 0.6209±0.0079/1.3503±0.0430 & 0.7215±0.0177/0.8431±0.0504 & 0.0835±0.0059/16.1396±0.5382 & 0.5406±0.0002/1.3186±0.0001 & 0.1217±0.0004/10.4389±0.0400 & 0.6640±0.0048/1.0289±0.0105 & 0.1364±0.0005/7.3042±0.6543 \\
\cdashline{1-13}[2pt/2pt]
\rule{0pt}{10pt}
2 & 0.7759±0.0017/0.7152±0.0054 & 0.5468±0.0197/1.7391±0.0836 & 0.6244±0.0014/1.1060±0.0007 & 0.4629±0.0054/2.1756±0.1009 & 0.6600±0.0062/1.0195±0.0055 & 0.6084±0.0138/1.3487±0.0409 & 0.7759±0.0017/0.7152±0.0054 & 0.1235±0.0017/11.4578±0.3880 & 0.6244±0.0014/1.1060±0.0007 & 0.1090±0.0071/23.1788±2.2104 & 0.6600±0.0062/1.0195±0.0055 & 0.1324±0.0035/4.2755±0.1208 \\
\cdashline{1-13}[2pt/2pt]
\rule{0pt}{10pt}
3 & 0.8014±0.0032/0.6448±0.0075 & 0.6420±0.0173/1.2342±0.1017 & 0.6835±0.0034/0.9741±0.0094 & 0.5093±0.0092/1.9345±0.0907 & 0.6798±0.0071/1.0373±0.0044 & 0.6179±0.0123/1.2980±0.0441 & 0.8014±0.0032/0.6448±0.0075 & 0.1019±0.0105/7.4709±0.2029 & 0.6835±0.0034/0.9741±0.0094 & 0.1013±0.0071/18.3295±1.9440 & 0.6798±0.0071/1.0373±0.0044 & 0.0970±0.0020/3.5487±0.0273 \\
\cdashline{1-13}[2pt/2pt]
\rule{0pt}{10pt}
4 & 0.8044±0.0025/0.6360±0.0016 & 0.6132±0.0074/1.3007±0.0519 & 0.7106±0.0002/0.8630±0.0045 & 0.5538±0.0154/1.8956±0.0235 & 0.6663±0.0049/1.0347±0.0146 & 0.6224±0.0179/1.2928±0.0583 & 0.8044±0.0025/0.6360±0.0016 & 0.1221±0.0155/5.0496±0.2744 & 0.7106±0.0002/0.8630±0.0045 & 0.1120±0.0020/12.3750±0.3344 & 0.6663±0.0049/1.0347±0.0146 & 0.1518±0.0097/2.8685±0.1497 \\
\cdashline{1-13}[2pt/2pt]
\rule{0pt}{10pt}
5 & 0.7923±0.0019/0.6858±0.0013 & 0.6199±0.0116/1.2408±0.0318 & 0.7343±0.0038/0.8023±0.0005 & 0.5452±0.0009/1.7285±0.0305 & 0.6572±0.0013/1.0551±0.0306 & 0.5737±0.0258/1.4265±0.0865 & 0.7923±0.0019/0.6858±0.0013 & 0.1492±0.0510/2.7731±0.5212 & 0.7343±0.0038/0.8023±0.0005 & 0.1103±0.0001/10.6789±0.2916 & 0.6572±0.0013/1.0551±0.0306 & 0.1186±0.0001/3.1417±0.0297 \\
\bottomrule[1.5pt]
\end{tabular}}
\label{tab: SITS_new_new}
\end{table*}

\subsection{Main Experimental Results}
In our main experiments, for each type of source model structure, we deepen the source model and then train the target model with different structures of input transformation layer. We verify our theoretical prediction by analyzing 
the loss of the source model and the loss of the target model, which is presented below.

\paragraph{Eigenvalue Spectrum of NTK}
We first compute the source model NTK ($K_S$). The experimental results suggest that, for all source model structures (CNN, VGG, ResNet), the minimum eigenvalue $\lambda_{\min}[K_S]$ increase as we deepen the source model (see Fig.~\ref{fig: NTKspectrum}).
According to Corollary~\ref{cor: spectrum} and Corollary~\ref{cor: spectrum_B}, we can predict that both loss of the source model and loss of the target model should decrease as the layer of the source model is deepen.


\paragraph{Source Loss v.s. Target Loss}
We examine the performance of the source model and the performance of the target model in Table~\ref{tab: SITC} and Table~\ref{tab: SITS}. The experimental results suggest that, for VGG and ResNet, both the loss of the source model and the loss of the target model decrease as we deepen the source model, which align with our theoretical prediction. As for CNN, the relation between the source loss and the target loss seems to deviate our theoretical prediction. We believe this deviation is because our Assumption~\ref{cor: spectrum_B} do not holds for CNN. We will justify our conjecture later.



\begin{figure}[h]
    \centering
    \scalebox{1}{
    \begin{minipage}{\linewidth}
    \begin{minipage}{0.3\linewidth}
        \centerline{\includegraphics[width=\textwidth,height=0.7\textwidth]{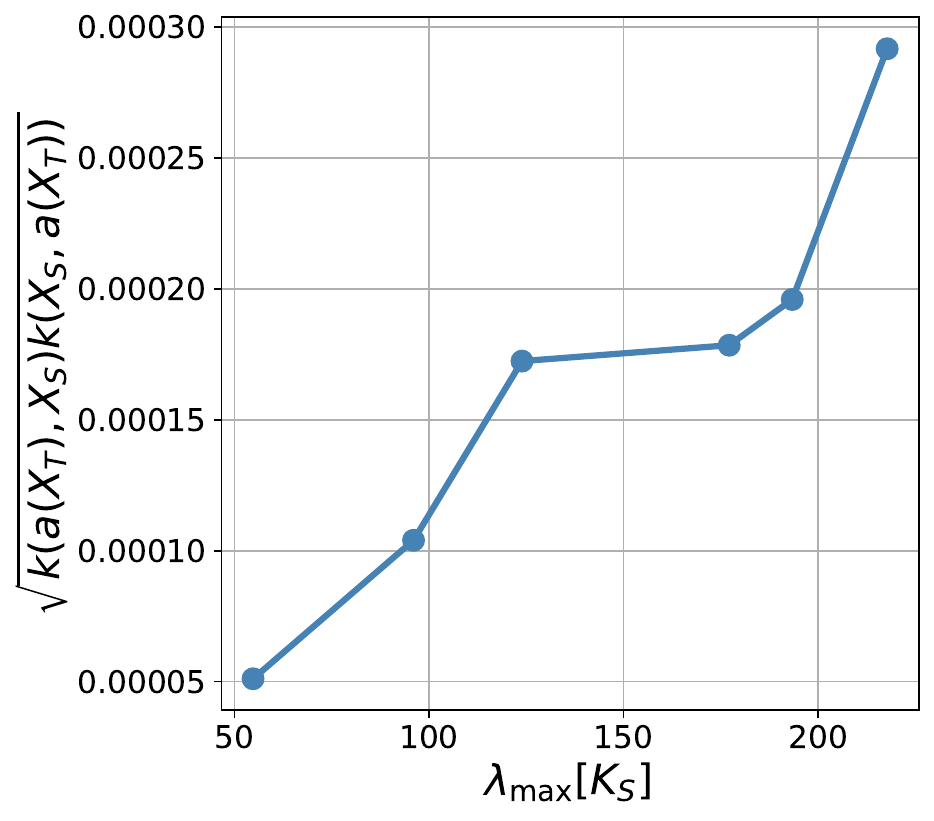}}
        \center{\footnotesize(a)VGG~T:C (FC)}
    \end{minipage}
    \hfill
    \begin{minipage}{0.3\linewidth}
        \centerline{\includegraphics[width=\textwidth,height=0.7\textwidth]{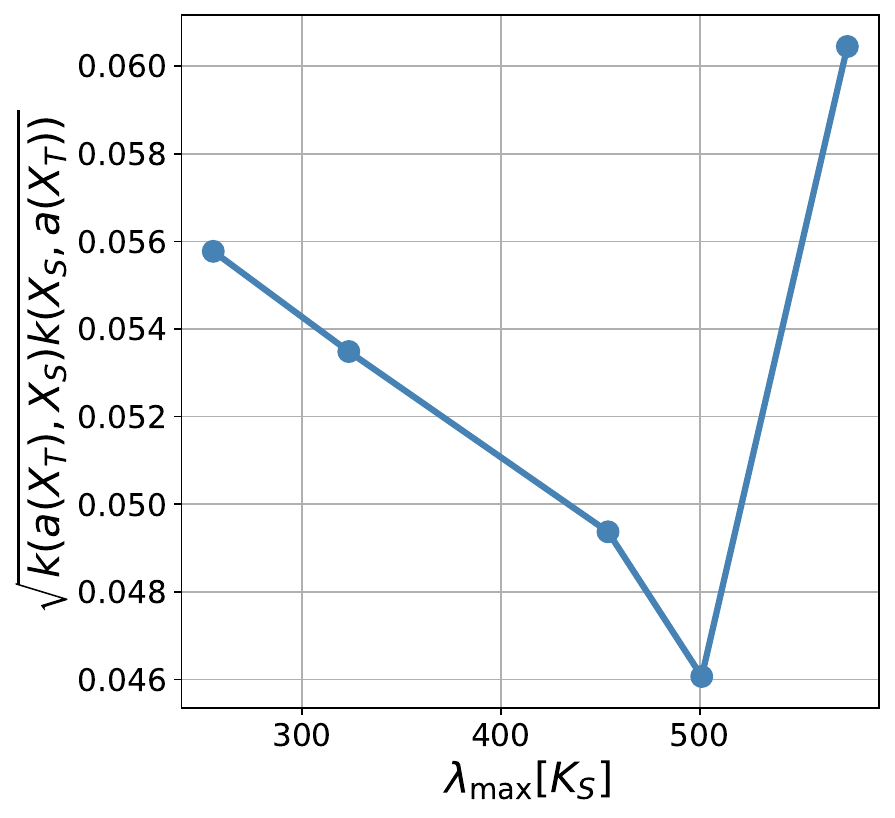}}
        \center{\footnotesize(b)CNN~T:C (FC)}
    \end{minipage}
    \hfill
    \begin{minipage}{0.3\linewidth}
        \centerline{\includegraphics[width=\textwidth,height=0.7\textwidth]{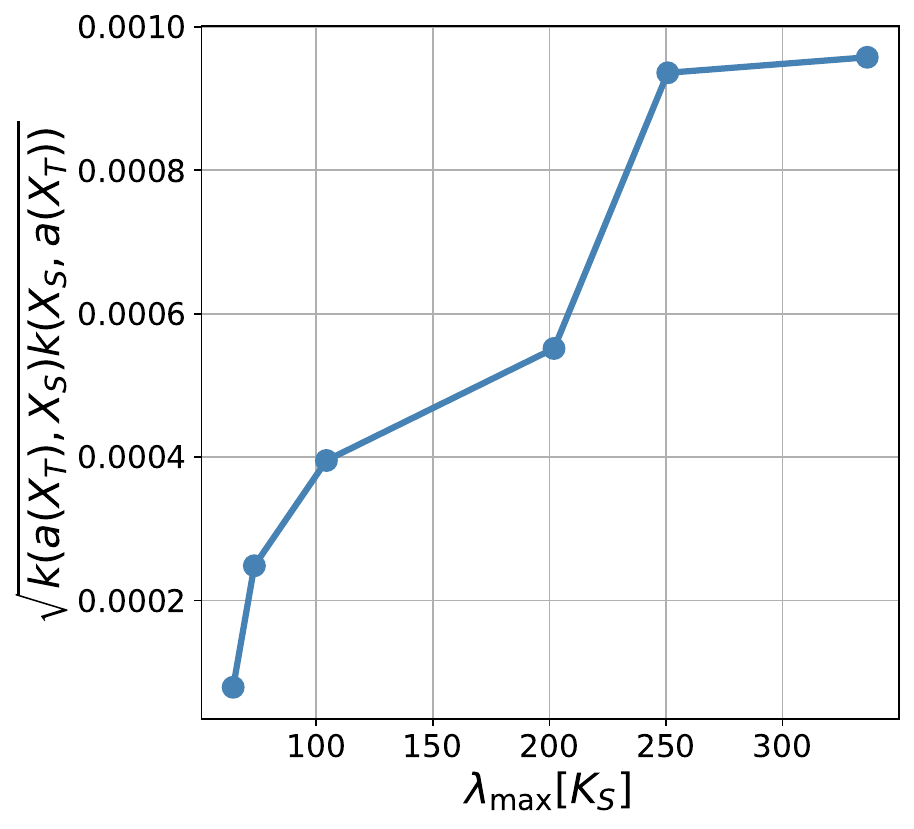}}
        \center{\footnotesize(c)RES~T:C (FC)}
    \end{minipage}
    \vspace{2pt}
    
    \begin{minipage}{0.3\linewidth}
        \centerline{\includegraphics[width=\textwidth,height=0.7\textwidth]{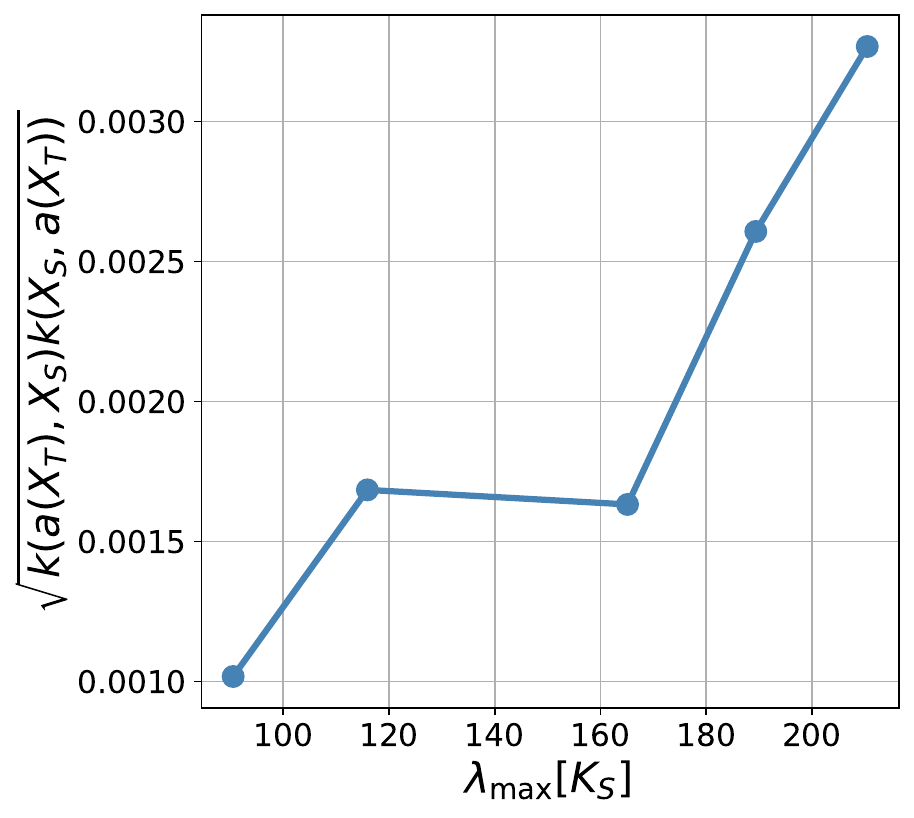}}
        \center{\footnotesize(d)VGG~T:C (VP)}
    \end{minipage}
    \hfill
    \begin{minipage}{0.3\linewidth}
        \centerline{\includegraphics[width=\textwidth,height=0.7\textwidth]{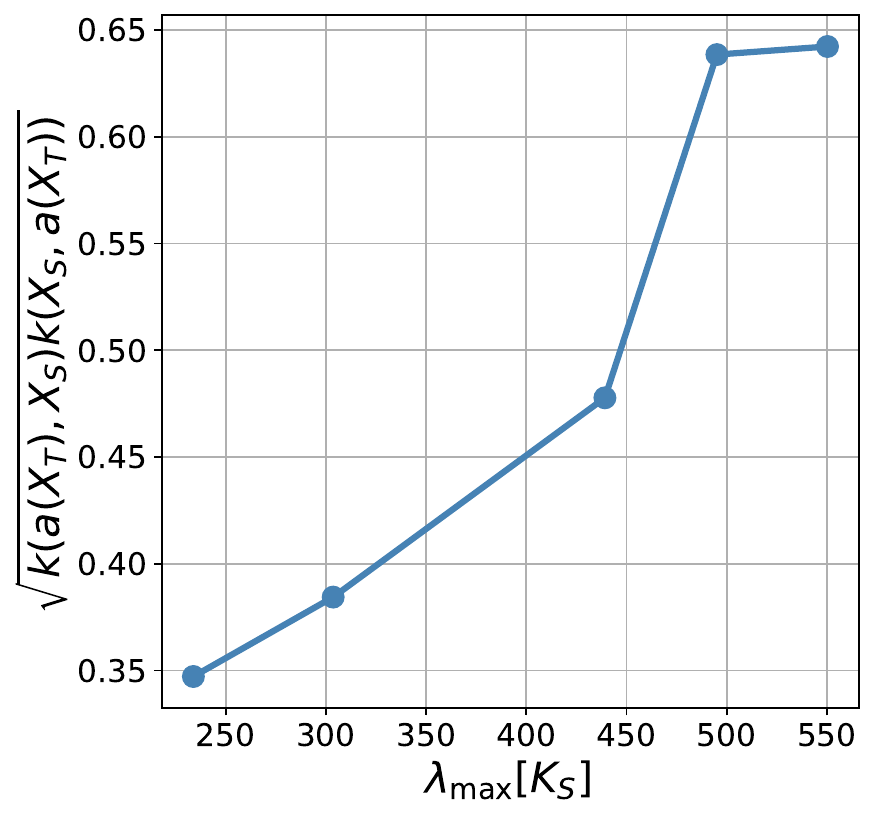}}
        \center{\footnotesize(e)CNN~T:C (VP)}
    \end{minipage}
    \hfill
    \begin{minipage}{0.3\linewidth}
        \centerline{\includegraphics[width=\textwidth,height=0.7\textwidth]{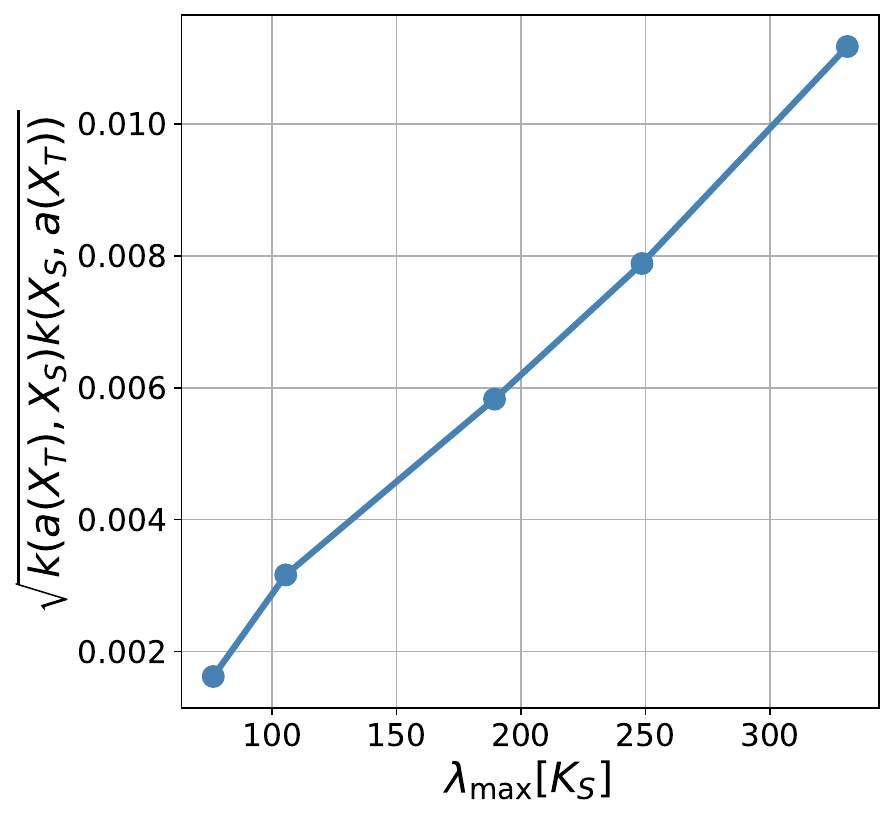}}
        \center{\footnotesize(f)RES~T:C (VP)}
    \end{minipage}
    \vspace{2pt}
    
    \begin{minipage}{0.3\linewidth}
        \centerline{\includegraphics[width=\textwidth,height=0.7\textwidth]{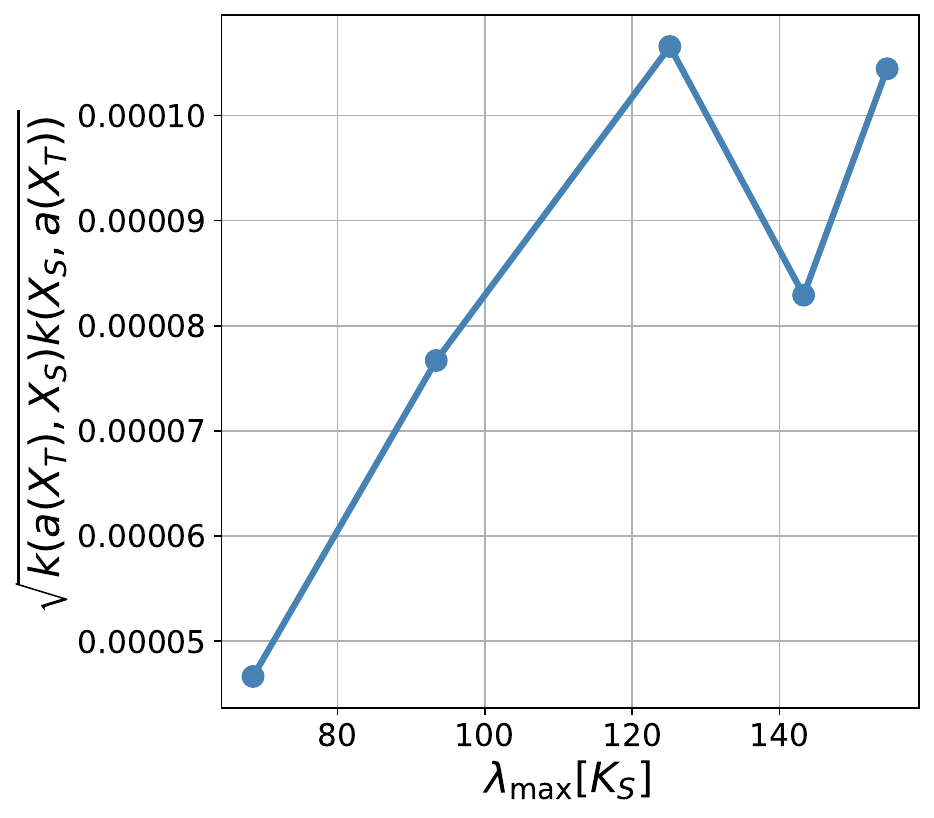}}
        \center{\footnotesize(g)VGG~T:SV (FC)}
    \end{minipage}
    \hfill
    \begin{minipage}{0.3\linewidth}
        \centerline{\includegraphics[width=\textwidth,height=0.7\textwidth]{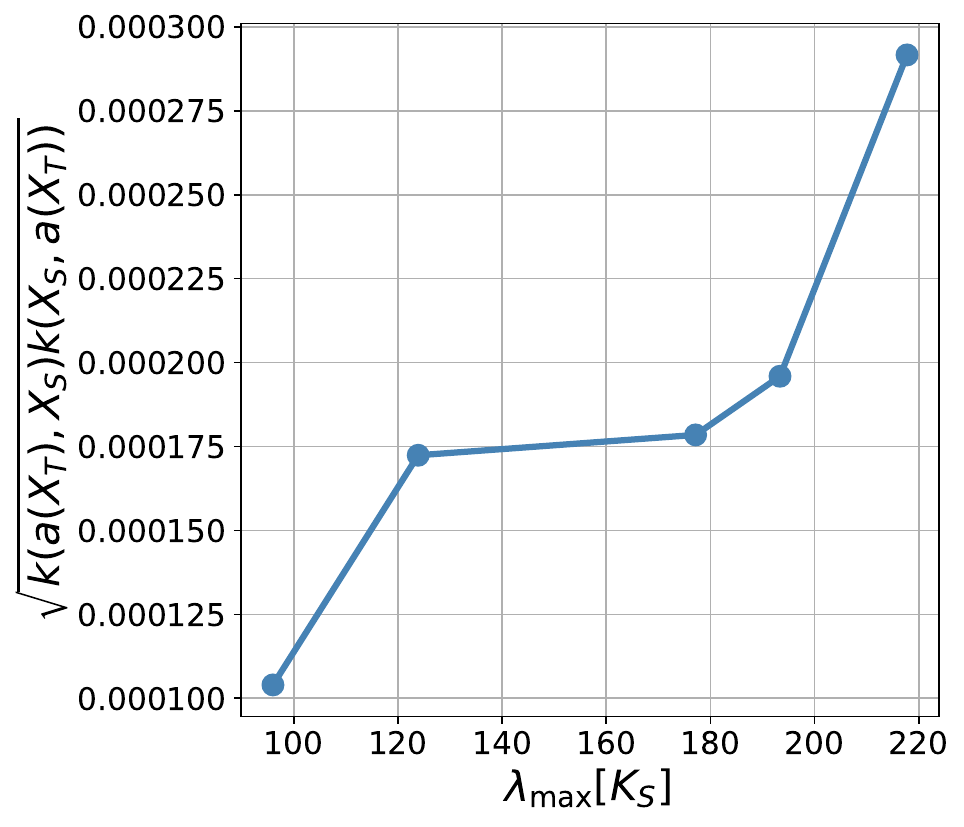}}
        \center{\footnotesize(h)CNN~T:SV (FC)}
    \end{minipage}
    \hfill
    \begin{minipage}{0.3\linewidth}
        \centerline{\includegraphics[width=\textwidth,height=0.7\textwidth]{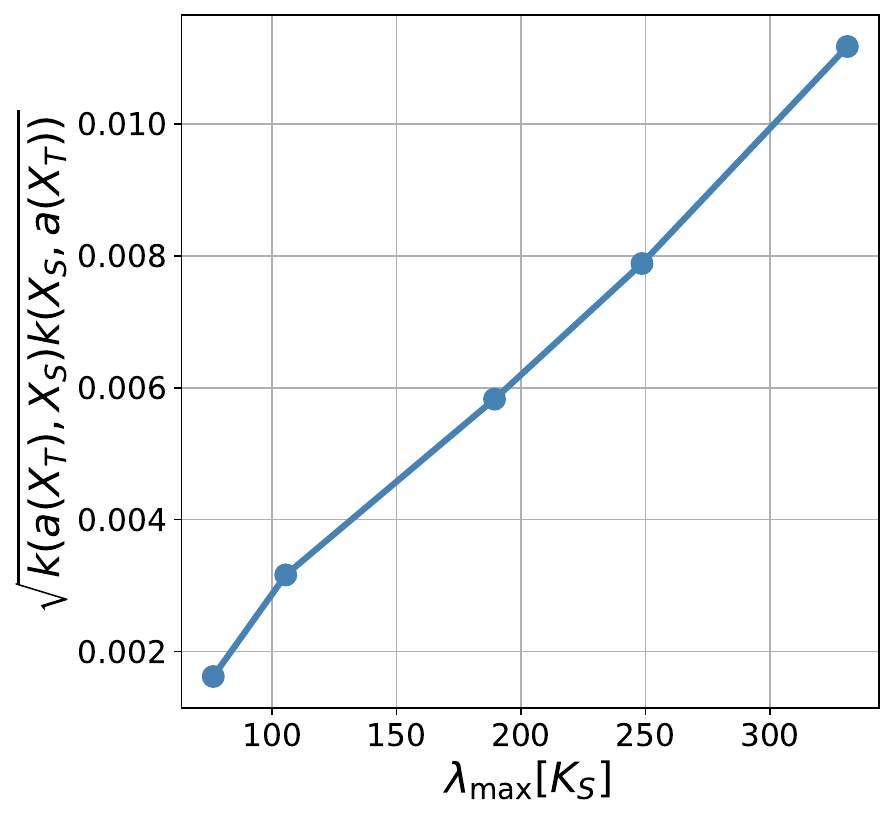}}
        \center{\footnotesize(i)RES~T:SV (FC)}
    \end{minipage}
    \vspace{2pt}

    \begin{minipage}{0.3\linewidth}
        \centerline{\includegraphics[width=\textwidth,height=0.7\textwidth]{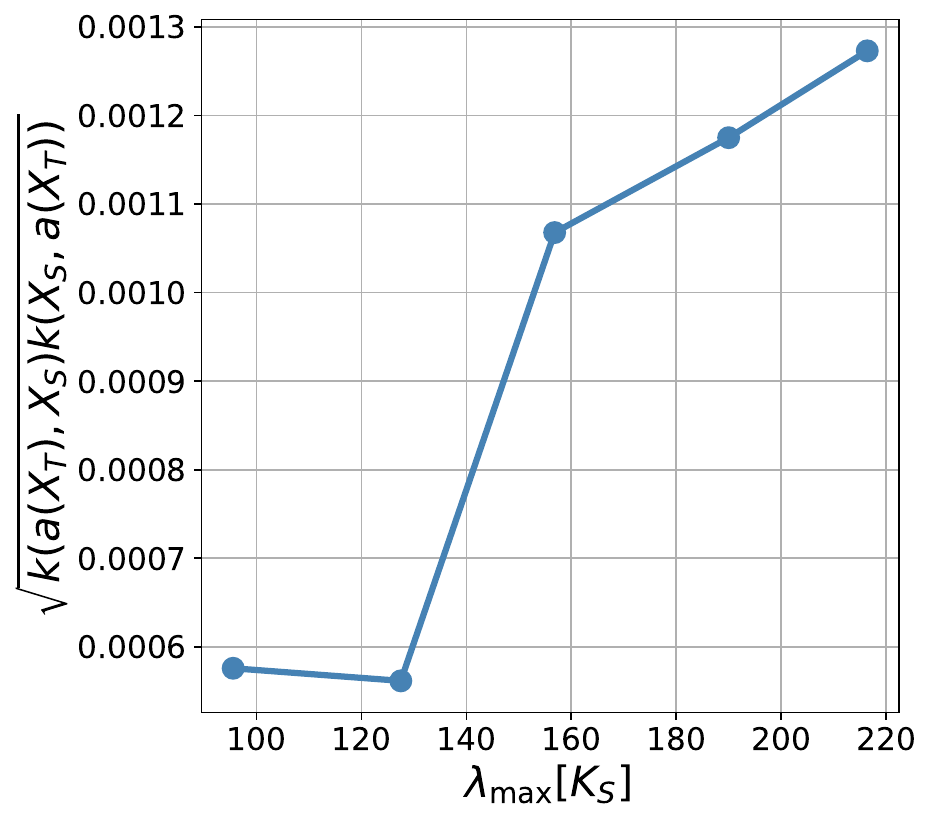}}
        \center{\footnotesize(j)VGG~T:SV (VP)}
    \end{minipage}
    \hfill
    \begin{minipage}{0.3\linewidth}
        \centerline{\includegraphics[width=\textwidth,height=0.7\textwidth]{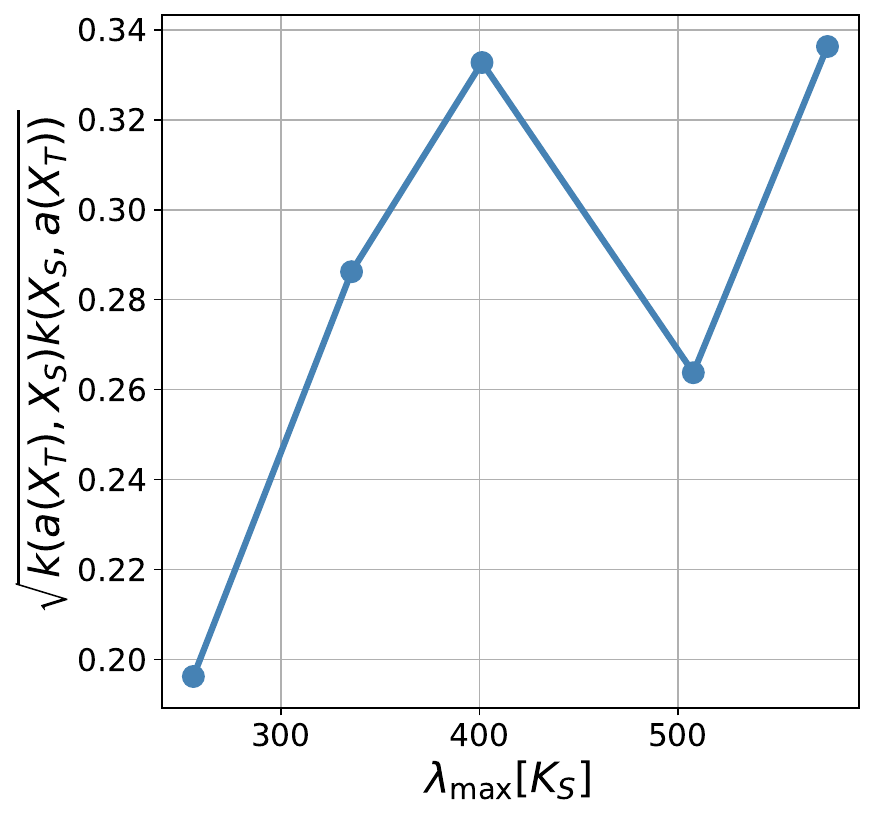}}
        \center{\footnotesize(k)CNN~T:SV (VP)}
    \end{minipage}
    \hfill
    \begin{minipage}{0.3\linewidth}
        \centerline{\includegraphics[width=\textwidth,height=0.7\textwidth]{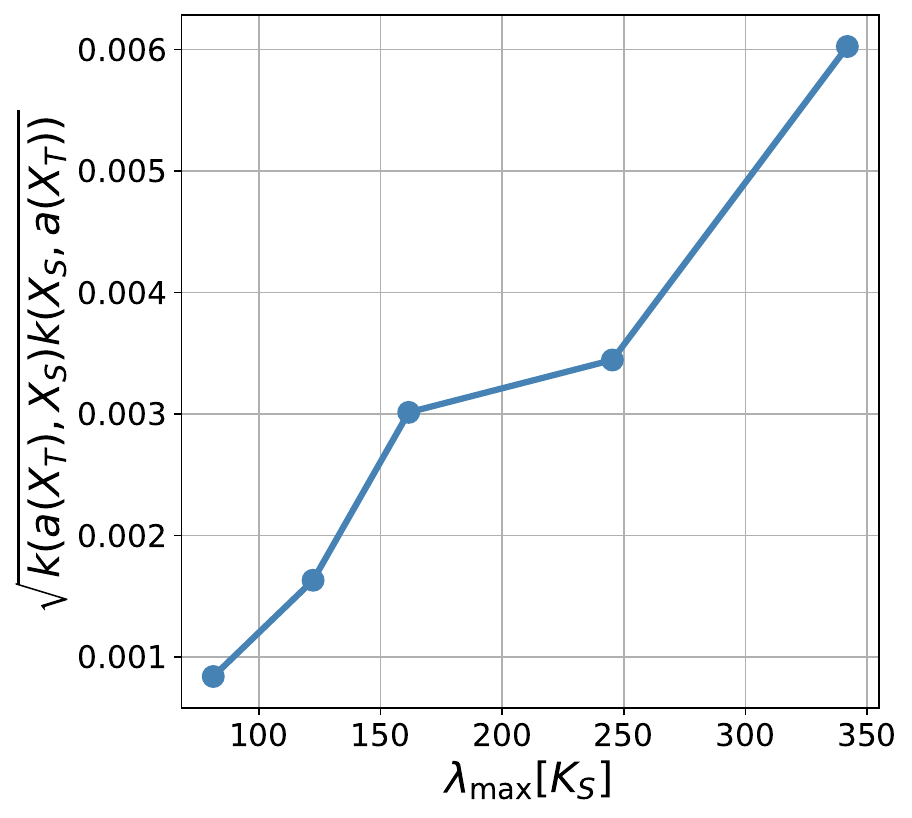}}
        \center{\footnotesize(l)RES~T:SV (VP)}
    \end{minipage}
    \end{minipage}
    }
    \caption{Assumption Justification} 
    \label{fig: assumption justification}
    \vspace{-10pt}
\end{figure}

\paragraph{Assumption Justification}
To validate Assumption~\ref{assump: cor_2}, we conducted empirical analysis comparing $\sqrt{\lambda_{min}[k(a(X_T), X_S)k(X_S, a(X_T))]}$ versus $\lambda_{\max}[K_S]$ across different model architectures. The results, presented in Fig~\ref{fig: assumption justification}, reveal distinct patterns:

For VGG and ResNet architectures, $\sqrt{\lambda_{min}[k(a(X_T), X_S)k(X_S, a(X_T))]}$ demonstrates first-order growth with respect to $\lambda_{\max}[K_S]$, exhibiting a clear linear relationship. This behavior aligns with our theoretical expectations under Assumption~\ref{assump: cor_2}.

However, CNN architectures show deviation from this pattern. Specifically, in subfigures (b), (h), and (k), we observe that when $\lambda_{min}[K_S]$ is small, $\sqrt{\lambda_{min}[k(a(X_T), X_S)k(X_S, a(X_T))]}$ exhibits sub-linear growth relative to $\lambda_{\max}[K_S]$, failing to maintain the first-order relationship predicted by Assumption~\ref{assump: cor_2}. This violation of our assumption provides a theoretical explanation for the discrepancy observed between source and target loss behaviors in CNN experiments (Tables~\ref{tab: SITC} and \ref{tab: SITS}).

\subsection{Other Experiments}
In addition to these baseline experiments with standard architectures, we also conducted extensive experiments using CLIP as a feature extractor to validate our theoretical framework at scale. These experiments follow a similar methodology while incorporating pre-trained representations from foundation models. The complete experimental setup and results are detailed in Section~\ref{sec: experiments} in appdendix.

\vspace{-3pt}
\section{Conclusion}
We first study the mechanism of MR through the NTK framework. Our theoretical results manifest that the eigenvalue spectrum of the source model NTK controls both performance of the source model and the performance of the target model. This explains the phenomenon that the success of MR usually depends on the success of the source model, which is found in \citep{tsao2023autovp, li2023exploring}. We conducted a series of numerical experiments. The experimental results further validate our theoretical prediction. In summary, our studies provide a theoretical perspective to analyze and understand Model Reprogramming (MR).

\clearpage
\newpage

\section*{Impact Statement}
This paper aims to deepen the understanding of Model Reprogramming (MR) through theoretical analysis. To the best of our knowledge, this is the first work to provide a formal theoretical foundation for MR. We believe our findings offer valuable insights that can guide future advancements in MR.

\bibliographystyle{abbrvnat}
\bibliography{ICML}

\newpage
\appendix
\onecolumn
\section{Appendix}\label{sec: Appendix}
\subsection{Notation Table}\label{sec: Notation Table}
The notations used in this paper are presented in Table~\ref{table:notations}.

\begin{table}[!ht]
    \centering
    \begin{small}
    \begin{tabular}{p{1in}p{3.25in}}
        \multicolumn{1}{l}{Notations} & \multicolumn{1}{c}{Descriptions}\\
        \midrule
        $\mathcal{X}$ & feature space $\subset\mathbb{R}^d$\\
        $\mathcal{Y}$ & label space $\subset\mathbb{R}^c$\\
        $x$ & feature\\
        $y$ & label\\
        $\mathfrak{D}$ & probability distribution distributed in $\mathcal{X}\times\mathcal{Y}$\\
        $D$ & dataset sampled form distribution $\mathfrak{D}$ with sample size $N$.\\
        $(X, Y)$ & matrix form of dataset $D$ where $X$ is a $\mathbb{R}^{N\times d}$ matrix and $Y$ is a $\mathbb{R}^{N\times c}$ matrix.\\
        
        $\mathcal{X}_S$ & feature space of source distribution $\subset\mathbb{R}^{d_S}$\\
        $\mathcal{Y}_S$ & label space of source distribution $\subset\mathbb{R}^{c_S}$\\
        $x_s$ & feature of source distribution\\
        $y_s$ & label of source distribution\\
        $\mathfrak{D}_S$ & source distribution distributed in $\mathcal{X}_S\times\mathcal{Y}_S$\\
        $D_S$ & source dataset sampled form source distribution $\mathfrak{D}_S$ with sample size $N_S$.\\
        $(X_S, Y_S)$ & matrix form of source dataset $D_S$ where $X_S$ is a $\mathbb{R}^{N_S\times d_S}$ matrix and $Y_S$ is a $\mathbb{R}^{N_S\times c_S}$ matrix.\\

        $\mathcal{X}_T$ & feature space of target distribution $\subset\mathbb{R}^{d_T}$\\
        $\mathcal{Y}_T$ & label space of target distribution $\subset\mathbb{R}^{c_T}$\\
        $x_t$ & feature of target distribution\\
        $y_t$ & label of target distribution\\
        $\mathfrak{D}_T$ & target distribution distributed in $\mathcal{X}_T\times\mathcal{Y}_T$\\
        $D_T$ & target dataset sampled form target distribution $\mathfrak{D}_T$ with sample size $N_T$.\\
        $(X_T, Y_T)$ & matrix form of target dataset $D_T$ where $X_T$ is a $\mathbb{R}^{N_T\times d_T}$ matrix and $Y_T$ is a $\mathbb{R}^{N_T\times c_T}$ matrix.\\

        $f$ & model mapping from $\mathbb{R}^d$ to $\mathbb{R}^c$\\
        $\mathcal{H}$ & hypothesis class of $f$\\

        $f_S$ & source model mapping from $\mathbb{R}^{d_S}$ to $\mathbb{R}^{c_S}$\\
        $\mathcal{H}_S$ & hypothesis class of $f_S$\\

        $f_T$ & target model mapping from $\mathbb{R}^{d_T}$ to $\mathbb{R}^{c_T}$\\
        $\mathcal{H}_T$ & hypothesis class of $f_T$\\

        $a$ & input transformation layer mapping from $\mathbb{R}^{d_T}$ to $\mathbb{R}^{d_S}$\\
        $\mathcal{H}_A$ & hypothesis class of $a$\\

        $b$ & output mapping from $\mathbb{R}^{c_S}$ to $\mathbb{R}^{c_T}$\\
        $\mathcal{H}_B$ & hypothesis class of $b$\\

        $k(\cdot, \cdot)$ & kernel defined in the Definition~\ref{def:kernel}\\
        $\mathcal{H}_k$ & reproducing kernel Hilbert space induced by kernel $k$, which is defined in the Definition~\ref{def:rkhs}\\

        $\hat\Theta(\cdot, \cdot)$ & NTK of $\mathcal{H}$ defined in Definition~\ref{def: NTK}\\
        $\Theta(\cdot, \cdot)$ & kernel induced by NTK, which is defined in Equation~\ref{eq: NTK->kernel}\\

        $\hat\Theta_T(\cdot, \cdot)$ & target NTK of $\mathcal{H}_T$ defined in Definition~\ref{def: NTK}\\
        $\Theta_T(\cdot, \cdot)$ & target kernel induced by target NTK, which is defined in Equation~\ref{eq: NTK->kernel}\\
        $\mathcal{H}_{\Theta_T}$ & reproducing kernel Hilbert space induced by kernel $\Theta_T$, which is defined in the Definition~\ref{def:rkhs}\\

        $\lambda_i(\cdot)$ & operator to output the $i$-th large eigenvalue\\
        $\lambda_{\text{max}}(\cdot)$ & operator to output the maximum eigenvalue\\
        $\lambda_{\text{min}}(\cdot)$ & operator to output the minimum eigenvalue\\
        \midrule
    \end{tabular}
    \end{small}
    \caption{Notation Table}
    \label{table:notations}
\end{table}

\subsection{Generalization Gap of MR}\label{sec: Generalization Gap of MR}
Let $\mathcal{G} = \{g : (x,y)\mapsto \|f(x)-y\|^2_2~|~ f\in\mathcal{H}_{\Theta_T}^{c_T}\}$, where $\mathcal{H}_{\Theta_T}$ is the RKHS corresponding to the kernel $\Theta_T$, $B>0$ be a parameter to bound the $\mathcal{H}_{\Theta_T}$ and $T>0$ be a parameter to describe the training length. 

\begin{assumption}\label{assump: general} 
To derive an upper bound of the generalization gap, we consider the following assumptions:
\begin{enumerate}
        \item $\|(x, y) - (x', y')\|_2 \leq\Gamma_{\mathfrak{D}_T},~\forall (x, y), (x', y')\sim\mathfrak{D}_T$,
        \item $\|\nabla_{(x, y)} g(x, y)\|_2 \leq L_{\mathfrak{D}_T},~\forall(x, y)\sim\mathfrak{D}_T$,
        \item $\|\nabla_{f} g(x, y)\|\leq\rho,~\forall      f\in\mathcal{H}_{\Theta_T}^{c_T}$,
        \item $\frac{T\cdot\rho^2}{N_T^2}\sum_{(x_t, y_t)\sim D_T \atop ({x'}_t, {y'}_t)\sim D_T} |\Theta_T (x_t, x_t')| \leq B^2$.
\end{enumerate}
\end{assumption}
By refining existing results (Theorem 3 in \citep{chung2024rethinking} and Eq.~4 in \citep{chen2024analyzing}), we can derive the following upper bound. 

\begin{theorem}[Upper bound of generalization gap]\label{thm:gen}
Given a $N_T$-sample target dataset $D_T$, sampled from the target distribution $\mathfrak{D}_T$. Suppose that Assumption~\ref{assump: general} holds, then the generalization gap defined in Eq.~\ref{eq: framework} can be bounded as follows with probability at least $1-\delta$:
\begin{small}
\begin{align}
    &\mathbb{E}_{(x_t,y_t)\sim\mathfrak{D}_T}[g(x_t, y_t)] - \sum_{(x_i, y_i)\in D_T} \frac{g(x_i, y_i)}{N_T} \nonumber\\
    &\leq 
    \frac{2 \rho B \sqrt{T}}{N_T} 
    \sum_{(x_t, y_t)\in D_T \atop (x_t', y_t')\in D_T} |\Theta_T (x_t, x'_t)|
    \nonumber\\
    &+ 
    3 L_{\mathfrak{D}_T} \Gamma_{\mathfrak{D}_T} \sqrt{\frac{\log{\frac{2}{\delta}}}{2N_T}}\label{eq:gengap},~\forall g\in\mathcal{G}.
\end{align}
\end{small}
\end{theorem}
The proof of Theorem~\ref{thm:gen} can be found in Appendix~\ref{sec: Theorem 2 and Its Proof}. It is important to note that the second term in Eq.~\ref{eq:gengap} reflects the intrinsic properties of the target distribution $\mathfrak{D}_T$ (specifically, $L_{\mathfrak{D}_T}$ and $\Gamma_{\mathfrak{D}_T}$), which are typically considered constants in most scenarios. Consequently, the generalization gap is primarily influenced by the first term in Eq.~\ref{eq:gengap}, which depends on the kernel matrix $\Theta_T (X_T, X_T)$. As $\|\Theta_T(X_T, X_T)\|_F$ approaches zero, the sum $\sum_{(x_t, y_t)\in D_T \atop (x_t', y_t')\in D_T} |\Theta_T (x_t, x'_y)|$ also converges to zero. Therefore, Theorem~\ref{thm:gen} suggests that reducing the Frobenius norm of the kernel matrix $\Theta_T (X_T, X_T)$ will result in a smaller generalization gap. However, in general, $\Gamma_{\mathfrak{D}_T}$ is supposed as a very large number. Hence, based on Eq.~\ref{eq:gengap}, the generalization gap could be insensitive when we slightly vary the NTK $\Theta_T(X_T, X_T)$.
\clearpage

\subsection{Theorem 1 and Its Proof}\label{sec: Theorem 1 and Its Proof}
\begin{customthm}{1}[Bound of Empirical Risk]\label{custhm: empirical risk}
Denote the eigenvalue of the kernel matrix $\Theta_T(X_T, X_T)$ as $\{ \lambda_i \}_{i=1}^{N_T}$ where $\lambda_i \geq \lambda_j$ for any $i < j$. The empirical risk of $\mathcal{L}_{\text{ER}}$ can be bounded as
\begin{equation}
\frac{1}{N_T}[1 - \frac{\lambda_{1}}{\sigma + \lambda_{1}}]\cdot\|Y_T\|_2^2
\leq 
\mathcal{L}_{\text{ER}} 
\leq 
\frac{1}{N_T}[1 - \frac{\lambda_{N_T}}{\sigma + \lambda_{N_T}}]\cdot\|Y_T\|_2^2.
\end{equation}
\end{customthm}
\begin{proof}
At first, by Definition~\ref{def:kernel}, we know that $\Theta_T(X_T, X_T)$ is postive semi-definite, which implies that $\Theta_T(X_T, X_T)$ can be diagonalized by some orthonormal system. Namely, there exists an unitary $U$ such that
\begin{equation}\label{eq: diagonal for ER}
    \Theta_T(X_T, X_T) = U \Sigma U^T
\end{equation}
where $\Sigma$ is the diagonal matrix whose diagonal components are eigenvalue of $\Theta_T(X_T, X_T)$. Then, by Equation~\ref{eq: diagonal for ER} and Equation~\ref{eq: def of ER}, we have
\begin{align}
    \mathcal{L}_{\text{ER}} 
    &=
    \frac{1}{N_T} 
    \|\{I -\Theta_T (X_T, X_T) [\Theta_T (X_T, X_T) + \sigma I]^{-1}\} Y_T \|_2^2\\
    &=
    \frac{1}{N_T} 
    \|\{I -  U \Sigma U^T [U \Sigma U^T + \sigma I]^{-1}\} Y_T \|_2^2\\
    &=
    \frac{1}{N_T} 
    \|\{I -  U \Sigma U^T [U \Sigma U^T + \sigma U I U^T]^{-1}\} Y_T \|_2^2\\
    &=
    \frac{1}{N_T} 
    \|\{I -  U \Sigma U^T U[ \Sigma + \sigma I]^{-1} U^T\} Y_T \|_2^2\\
    &=
    \frac{1}{N_T} 
    \|\{UIU^T -  U \Sigma[\Sigma + \sigma I]^{-1} U^T\} Y_T \|_2^2\\
    &=
    \frac{1}{N_T} 
    \|U\{I - \Sigma[ \Sigma + \sigma I]^{-1}\} U^T Y_T \|_2^2\\
    &=
    \frac{1}{N_T} 
    \|\{I - \Sigma[ \Sigma + \sigma I]^{-1}\} U^T Y_T \|_2^2\label{eq: proof of thm1_1}.
\end{align}
Clearly, Equation~\ref{eq: proof of thm1_1} can be bounded by the eigenvalue of $I - \Sigma[\Sigma+\sigma I]^{-1}$. We can obtain
\begin{align}
    \frac{1}{N_T} 
    \|\{I - \Sigma[ \Sigma + \sigma I]^{-1}\} U^T Y_T \|_2^2
    &\leq 
    \frac{1}{N_T}\lambda_{\text{max}}[I - \Sigma[ \Sigma + \sigma I]^{-1}] \|U^T Y_T\|_2^2.\nonumber\\
    &=
    \frac{1}{N_T}\lambda_{\text{max}}[I - \Sigma[ \Sigma + \sigma I]^{-1}] \| Y_T\|_2^2.\label{eq: proof of thm1_2}
\end{align}
and 
\begin{align}
    \frac{1}{N_T} 
    \|\{I - \Sigma[ \Sigma + \sigma I]^{-1}\} U^T Y_T \|_2^2
    &\geq 
    \frac{1}{N_T}\lambda_{\text{min}}[I - \Sigma[ \Sigma + \sigma I]^{-1}] \|U^T Y_T\|_2^2.\nonumber\\
    &=
    \frac{1}{N_T}\lambda_{\text{min}}[I - \Sigma[ \Sigma + \sigma I]^{-1}] \| Y_T\|_2^2.\label{eq: proof of thm1_3}
\end{align}
We also know that 
\[
\begin{cases}
    &\lambda_{\text{max}}[I - \Sigma[ \Sigma + \sigma I]^{-1}] = 1 - \frac{\lambda_{N_T}}{\sigma+\lambda_{N_T}}\\
    &\lambda_{\text{min}}[I - \Sigma[ \Sigma + \sigma I]^{-1}] = 1 - \frac{\lambda_{1}}{\sigma+\lambda_{1}}.
\end{cases}
\]
Hence, with Equation~\ref{eq: proof of thm1_2} and Equation~\ref{eq: proof of thm1_3}, we can derive
\begin{equation}
\frac{1}{N_T}[1 - \frac{\lambda_{1}}{\sigma + \lambda_{1}}]\cdot\|Y_T\|_2^2
\leq 
\mathcal{L}_{\text{ER}} 
\leq 
\frac{1}{N_T}[1 - \frac{\lambda_{N_T}}{\sigma + \lambda_{N_T}}]\cdot\|Y_T\|_2^2.
\end{equation}
which is what we want.
\end{proof}
\clearpage

\subsection{Proposition 1 and Its Proof}\label{sec: Proposition 1 and Its Proof}
\begin{customprop}{1}\label{cusprop: spectrum}
    Assume that the width of the target model $f_T: \mathbb{R}^{d_T}\rightarrow\mathbb{R}^{c_T}$ is sufficient large and hence $\hat\Theta_T(x, x') \rightarrow \Theta_T(x, x') I_{c_T}$, then the eigenvalue spectrum of $\hat\Theta_T (X_T, X_T)$ is equivalent to the eigenvalue spectrum of $\Theta_T(X_T, X_T)$.

    To be more specific, denote $\{\lambda_i\}_{i=1\sim N_T}$ as the eigenvalue spectrum of $\Theta_T(X_T, X_T)$, the eigenvalue spectrum of $\hat\Theta_T (X_T, X_T)$ will be $\{\lambda_i^j\}_{i=1\sim N_T \atop j=1\sim c_T}$ where $\lambda_i^j = \lambda_i$ for all $i$ and $j$.
\end{customprop}
\begin{proof}
Notice that $\hat\Theta_T(x, x') \rightarrow \Theta_T(x, x') I_{c_T}$ implies 
\begin{equation}
    \hat\Theta_T(X_T, X_T) = \Theta_T(X_T, X_T) \otimes I_{c_T}\label{eq: proof of prop1_1}
\end{equation}
where $\Theta_T(X_T, X_T)$ is a $N_T \times N_T$ matrix and $I_{c_T}$ is a $c_T \times c_T$ matrix, $\otimes$ is tensor product. Given a $n\times n$ matrix $A$ and a $m\times m$ matrix $B$, the eigenvalues of $A\otimes B$ will be
\begin{equation}
    \{\lambda_i (A\otimes B)~|~i=1\sim n\cdot m\} 
    =
    \{\lambda_i (A) \cdot \lambda_j (B)~|~i= 1\sim n,~j=1\sim m\}\label{eq: proof of prop1_2}
\end{equation}
where $\lambda_i(\cdot)$ is an operator outputting the $i$-th large eigenvalue. Hence, with Equation~\ref{eq: proof of prop1_1} and Equation~\ref{eq: proof of prop1_2}, we 
have
\begin{align}
     &\{\lambda_i (\Theta_T(X_T, X_T)\otimes I_{c_T})~|~i=1\sim N_T\cdot c_T\}\\
    &=
    \{\lambda_i (\Theta_T(X_T, X_T)) \cdot \lambda_j (I_{c_T})~|~i= 1\sim N_T,~j=1\sim c_T\}\\
    &=
    \{\lambda_i (\Theta_T(X_T, X_T)) \cdot 1~|~i= 1\sim N_T,~j=1\sim c_T\}
\end{align}
Thus, we can conclude that $\hat\Theta_T(X_T, X_T)$ and $\Theta_T(X_T, X_T)$ share the same eigenvalue spectrum.
\end{proof}

\subsection{Theorem 2 and Its Proof}\label{sec: Theorem 3 and Its Proof}
\begin{customassum}{1}\label{cusassump: ntk}
In order to clearly express the relation between $\Theta_T (X_T, X_T)$, target distribution $\mathfrak{D}_T$ and source distribution $\mathfrak{D}_S$. We consider the following assumptions:
\begin{enumerate}
        \item The source model $f_S$ can express some kernel model. Namely,
        \begin{equation}
            f_S (\cdot) = k(\cdot)[K_S + \sigma_S I] Y_S
        \end{equation}
        where the kernel matrix $K_S = k(X_S, X_S)$, $\sigma_S > 0$ is regularization parameter and the kernel $k(x, x') = \langle \Phi(x), \Phi(x')\rangle$ is induced by NTK.
        \item $\forall b\in\mathcal{H}_B$ is a $c_T \times c_S$ linear matrix and $b = [b^1, b^2, \dots, b^{c_T}]^T$ where $b^i \in\mathbb{R}^{c_S}$. To be more specific, the hypothesis class $\mathcal{H}_B = \{b~|~\mathbb{R}^{c_T \times c_S}\}$.
\end{enumerate}
\end{customassum}
\begin{customthm}{2}\label{custhm: spectrum}
Suppose the Assumption~\ref{cusassump: ntk} holds, the eigenvalue spectrum of the kernel matrix $\hat\Theta_T^A(X_T, X_T)$ can be bounded as follows
\begin{small}
\begin{equation}
    \lambda_i (\hat\Theta_T^A(X_T, X_T)) 
    \leq
    \lambda_{\text{max}}[\Theta_S^b] 
    \cdot\sup_{(x_t, y_t) \in D_T}\lambda_{\text{max}}[\hat\Theta_S^A(x_t, x_t)]
    \cdot \lambda_{\text{max}}[\hat\Theta_A(X_T, X_T)]
\end{equation}
\end{small}
and
\begin{small}
\begin{equation}
    \lambda_i (\hat\Theta_T^A(X_T, X_T)) 
    \geq
    \lambda_{\text{min}}[\Theta_S^b] 
    \cdot\inf_{(x_t, y_t) \in D_T}\lambda_{\text{min}}[\hat\Theta_S^A(x_t, x_t)]
    \cdot \lambda_{\text{min}}[\hat\Theta_A(X_T, X_T)]
\end{equation}
\end{small}
where $\lambda_i(\cdot)$ is the operator to output the $i$-th large eigenvalue,
$\hat\Theta^A(x, x') = \nabla_{\theta_A} a(x) [\nabla_{\theta_A} a(x')]^T$, $\hat\Theta^A_S(x, x') = \nabla_a \Phi(a(x)) [\nabla_a \Phi(a(x'))]^T$, $\Theta_S^b = b Y_S^T [K_S + \sigma_S I]^{-1} K_S [K_S + \sigma_S I]^{-1} Y_S b^T$.
\end{customthm}
\begin{proof}
Equation~\ref{eq: ntk_mr} tells us
\begin{small}
\begin{align}
    &\hat\Theta_T^A (x, x') =\nonumber\\ 
    &b Y_S^T [K_S + \sigma_S I]^{-1} \Phi(X_S)^T \nabla_{a} \Phi(a(x))
    \nabla_{\theta_A} a(x)
    \{b Y_S^T [K_S + \sigma_S I]^{-1} \Phi(X_S)^T \nabla_{a} \Phi(a(x))
    \nabla_{\theta_A} a(x)\}^T
\end{align}
\end{small}
which implies that 
\begin{small}
\begin{align}
    &\hat\Theta_T^A (X_T, X_T) = Q S \hat\Theta_A(X_T, X_T) S^T Q^T
\end{align}
\end{small}
where $\hat\Theta^A(x, x') = \nabla_{\theta_A} a(x) [\nabla_{\theta_A} a(x')]^T$ and
\begin{small}

\begin{equation}
    Q =
    \underbrace{
    \begin{pmatrix}
    b Y_S^T [K_S + \sigma_S I]^{-1} \Phi(X_S)^T &  &  &\\
     & b Y_S^T [K_S + \sigma_S I]^{-1} \Phi(X_S)^T &  &\\
     &  & \ddots & \\
      &  &  & b Y_S^T [K_S + \sigma_S I]^{-1} \Phi(X_S)^T
    \end{pmatrix}}_{N_T \text{ times}},
\end{equation}
\begin{equation}
    S =
    \underbrace{
    \begin{pmatrix}
    \nabla_{a} \Phi(a(x_1)) &  &  &\\
     & \nabla_{a} \Phi(a(x_2)) &  &\\
     &  & \ddots & \\
      &  &  & \nabla_{a} \Phi(a(x_{N_T}))
    \end{pmatrix}}_{N_T \text{ times}}.
\end{equation}
\end{small}
Then, by the definition of maximum eigenvalue, we can derive
\begin{align}
    \lambda_i (\hat\Theta_T^A(X_T, X_T)) 
    &\leq
    \lambda_{\text{max}} (\hat\Theta_T^A(X_T, X_T))\nonumber\\ 
    &=
    \sup_{\{x\in\mathbb{R}^{c_T}~|~\|x\|_2 =1\}}
    x^T Q S \hat\Theta_A(X_T, X_T) S^T Q^T x \\
    &\leq
    \sup_{\{x\in\mathbb{R}^{c_T}~|~\|x\|_2 =1\}}
    \lambda_{\text{max}}[\hat\Theta_A (X_T, X_T)] \cdot x^T Q S S^T Q^T x\\
    &=
    \lambda_{\text{max}}[\hat\Theta_A (X_T, X_T)]
    \cdot
    \sup_{\{x\in\mathbb{R}^{c_T}~|~\|x\|_2 =1\}}
    x^T Q S S^T Q^T x \\
    &\leq
    \lambda_{\text{max}}[\hat\Theta_A (X_T, X_T)]
    \cdot
    \sup_{\{x\in\mathbb{R}^{c_T}~|~\|x\|_2 =1\}}
    \lambda_{\text{max}} [S S^T]
    \cdot
    x^T Q Q^T x \\
    &=
    \lambda_{\text{max}}[\hat\Theta_A (X_T, X_T)]
    \cdot
    \lambda_{\text{max}} [S S^T]
    \cdot
    \sup_{\{x\in\mathbb{R}^{c_T}~|~\|x\|_2 =1\}}
    x^T Q Q^T x \\
    &\leq
    \lambda_{\text{max}}[\hat\Theta_A (X_T, X_T)]
    \cdot
    \lambda_{\text{max}} [S S^T]
    \cdot
    \sup_{\{x\in\mathbb{R}^{c_T}~|~\|x\|_2 =1\}}
    \lambda_{\text{max}} [Q Q^T] 
    \cdot
    \|x\|_2\\
    &=
    \lambda_{\text{max}}[\hat\Theta_A (X_T, X_T)]
    \cdot
    \lambda_{\text{max}} [S S^T]
    \cdot
    \lambda_{\text{max}} [Q Q^T]\label{eq: proof of thm3_0}
\end{align}
For $Q Q^T$, we have
\begin{small}
\begin{align}
    &Q Q^T =\\
    &\underbrace{
    \begin{pmatrix}
    b Y_S^T [K_S + \sigma_S I]^{-1} K_S \{b Y_S^T [K_S + \sigma_S I]^{-1}\}^T  &    \\
     & \ddots &  \\
      &   & b Y_S^T [K_S + \sigma_S I]^{-1} K_S \{b Y_S^T [K_S + \sigma_S I]^{-1}\}^T
    \end{pmatrix}}_{N_T \text{ times}}.\\
\end{align}
\end{small}
Hence,
\begin{align}
    \lambda_{\text{max}} [Q Q^T]
    &=
    \lambda_{\text{max}}[b Y_S^T [K_S + \sigma_S I]^{-1} K_S \{b Y_S^T [K_S + \sigma_S I]^{-1}\}^T]\\
    &=
    \lambda_{\text{max}}[\Theta_S^b]\label{eq: proof of thm3_1}.
\end{align}

For $S S^T$, we have
\begin{small}
\begin{align}
    &S S^T =\\
    &\underbrace{
    \begin{pmatrix}
    \nabla_{a} \Phi(a(x_1)) [\nabla_{a} \Phi(a(x_1))]^T  & &    \\
     & \nabla_{a} \Phi(a(x_1)) [\nabla_{a} \Phi(a(x_2))]^T & &  \\
      &   & \ddots & \\
      & & & \nabla_{a} \Phi(a(x_{N_T})) [\nabla_{a} \Phi(a(x_{N_T}))]^T
    \end{pmatrix}}_{N_T \text{ times}}
\end{align}
\end{small}
which implies that 
\begin{align}
    \lambda_{\text{max}}[SS^T] 
    &=
    \sup_{(x_t, y_t)\in D_T} \lambda_{\text{max}} \nabla_{a} \Phi(a(x_t)) [\nabla_{a} \Phi(a(x_t))]^T\\
    &=
    \sup_{(x_t, y_t)\in D_T} \lambda_{\text{max}} \nabla_{a} 
    \hat\Theta^A_S (x_t, x_t).\label{eq: proof of thm3_2}
\end{align}
Thus, combine Equation~\ref{eq: proof of thm3_0}, Equation~\ref{eq: proof of thm3_1} and Equation~\ref{eq: proof of thm3_2}, we can derive
\begin{equation}
    \lambda_i (\hat\Theta_T^A(X_T, X_T)) 
    \leq
    \lambda_{\text{max}}[\Theta_S^b] 
    \cdot\sup_{(x_t, y_t) \in D_T}\lambda_{\text{max}}[\hat\Theta_S^A(x_t, x_t)]
    \cdot \lambda_{\text{max}}[\hat\Theta_A(X_T, X_T)]
\end{equation}
which is what we want. 

On the other hand, by the definition of minimum eigenvalue, we can derive
\begin{align}
    \lambda_i (\hat\Theta_T^A(X_T, X_T)) 
    &\geq
    \lambda_{\text{min}} (\hat\Theta_T^A(X_T, X_T))\nonumber\\ 
    &=
    \inf_{\{x\in\mathbb{R}^{c_T}~|~\|x\|_2 =1\}}
    x^T Q S \hat\Theta_A(X_T, X_T) S^T Q^T x \\
    &\geq
    \inf_{\{x\in\mathbb{R}^{c_T}~|~\|x\|_2 =1\}}
    \lambda_{\text{min}}[\hat\Theta_A (X_T, X_T)] \cdot x^T Q S S^T Q^T x\\
    &=
    \lambda_{\text{min}}[\hat\Theta_A (X_T, X_T)]
    \cdot
    \inf_{\{x\in\mathbb{R}^{c_T}~|~\|x\|_2 =1\}}
    x^T Q S S^T Q^T x \\
    &\geq
    \lambda_{\text{min}}[\hat\Theta_A (X_T, X_T)]
    \cdot
    \inf_{\{x\in\mathbb{R}^{c_T}~|~\|x\|_2 =1\}}
    \lambda_{\text{min}} [S S^T]
    \cdot
    x^T Q Q^T x \\
    &=
    \lambda_{\text{min}}[\hat\Theta_A (X_T, X_T)]
    \cdot
    \lambda_{\text{min}} [S S^T]
    \cdot
    \inf_{\{x\in\mathbb{R}^{c_T}~|~\|x\|_2 =1\}}
    x^T Q Q^T x \\
    &\geq
    \lambda_{\text{min}}[\hat\Theta_A (X_T, X_T)]
    \cdot
    \lambda_{\text{min}} [S S^T]
    \cdot
    \inf_{\{x\in\mathbb{R}^{c_T}~|~\|x\|_2 =1\}}
    \lambda_{\text{min}} [Q Q^T] 
    \cdot
    \|x\|_2\\
    &=
    \lambda_{\text{min}}[\hat\Theta_A (X_T, X_T)]
    \cdot
    \lambda_{\text{min}} [S S^T]
    \cdot
    \lambda_{\text{min}} [Q Q^T]\label{eq: proof of thm3_3}.
\end{align}
With similar process, we can conclude that 
\begin{equation}
    \lambda_i (\hat\Theta_T^A(X_T, X_T)) 
    \geq
    \lambda_{\text{min}}[\Theta_S^b] 
    \cdot\inf_{(x_t, y_t) \in D_T}\lambda_{\text{min}}[\hat\Theta_S^A(x_t, x_t)]
    \cdot \lambda_{\text{min}}[\hat\Theta_A(X_T, X_T)]
\end{equation}
which is what we want.
\end{proof}

\subsection{Corollary 1 and Its Proof}\label{sec: Corollary 1 and Its Proof}
\begin{customassum}{2}\label{cusassump: cor}
Given a source dataset $D_S$ and a target dataset $D_T$, we suppose that there exists $c_A > 0 $ such that
\begin{equation}
    \lambda_{\text{min}}[\hat\Theta^A_S(x_t, x_t)] \geq c_A 
    \cdot 
    (\lambda_{\text{max}}[K_S] + \sigma_S)~,\forall (x_t, y_t)\in D_T.
\end{equation}
\end{customassum}
\begin{customcor}{1}\label{cuscor: spectrum}
Suppose Assumption~\ref{cusassump: ntk} and Assumption~\ref{cusassump: cor} hold, then we have
\begin{align}
    \lambda_i (\hat\Theta_T^A(X_T, X_T)) 
    \geq
    \lambda_{\text{min}}(b Y_S^T Y_S b^T) 
    \cdot
    c_A
    \cdot
    [\frac{\lambda_{\text{min}}[K_S]}{\lambda_{\text{min}}[K_S] + \sigma}]
    \cdot
    \lambda_{\text{min}}[\hat\Theta_A(X_T, X_T)].
\end{align}
\end{customcor}
\begin{proof}
By Theorem~\ref{custhm: spectrum}, we know that
\begin{equation}
    \lambda_i (\hat\Theta_T^A(X_T, X_T)) 
    \geq
    \lambda_{\text{min}}[\Theta_S^b] 
    \cdot\inf_{(x_t, y_t) \in D_T}\lambda_{\text{min}}[\hat\Theta_S^A(x_t, x_t)]
    \cdot \lambda_{\text{min}}[\hat\Theta_A(X_T, X_T)].
\end{equation}
Notice that we have assumed $\lambda_{\text{min}}[\hat\Theta^A_S(x_t, x_t)] \geq c_A \cdot (\lambda_{\text{max}}[K_S]+\sigma_S)$. Then, we can derive
\begin{align}
    \lambda_i (\hat\Theta_T^A(X_T, X_T)) 
    &\geq
    \lambda_{\text{min}}[\Theta_S^b] 
    \cdot\inf_{(x_t, y_t) \in D_T}\lambda_{\text{min}}
    [\hat\Theta_S^A(x_t, x_t)]
    \cdot \lambda_{\text{min}}[\hat\Theta_A(X_T, X_T)]\\
    &\geq
    \lambda_{\text{min}}[\Theta_S^b]
    \cdot c_A
    \cdot (\lambda_{\text{max}}[K_S] + \sigma_S)
    \cdot \lambda_{\text{min}}[\hat\Theta_A(X_T, X_T)].\label{eq: proof of cor1_1}
\end{align}
For $\lambda_{\text{min}}[\Theta^b_S]$, we can obtain
\begin{align}
    \lambda_{\text{min}}[\Theta^b_S]
    &=
    \lambda_{\text{min}}[b Y_S^T [K_S + \sigma_S I]^{-1} K_S \{b Y_S^T [K_S + \sigma_S I]^{-1}\}^T]\\
    &\geq
    \lambda_{\text{min}}[b Y_S^T Y_S b^T]
    \cdot
    \lambda_{\text{min}}
    [[K_S + \sigma_S I]^{-1} K_S [K_S + \sigma_S I]^{-1} ]\label{eq: proof of cor1_2}.
\end{align}
Since $K_S$ is positive semi-definite, we can express $K_S$ as $U \Sigma U^T$ where $U$ is some unitary matrix and $\Sigma$ is a diagonal matrix whose components are eigenvalues of $K_S$. Hence, we can derive
\begin{align}
    [K_S + \sigma_S I]^{-1} K_S
    &=
    [U\Sigma U^T + \sigma_S U I U^T]^{-1} 
    U\Sigma U^T\\
    &=
    U[\Sigma + \sigma_S I]^{-1} U^T
    U \Sigma U^T\\
    &=
    U[\Sigma + \sigma_S I]^{-1}
    \Sigma U^T.
\end{align}
Furthermore, 
\begin{align}
    \lambda_{\text{min}}[[K_S + \sigma_S I]^{-1} K_S [K_S + \sigma_S I]^{-1}]
    &\geq
    \lambda_{\text{min}}[[K_S + \sigma_S I]^{-1} K_S]
    \cdot
    \lambda_{\text{min}}[[K_S + \sigma_S I]^{-1}]\\
    &=
    \lambda_{\text{min}}[[\Sigma + \sigma_S I]^{-1}
    \Sigma] \cdot (\lambda_{\text{max}}[K_S] + \sigma_S)^{-1}\\
    &=
    \frac{\lambda_{\text{min}}[\Sigma]}{[\lambda_{\text{min}}[\Sigma] + \sigma_S]} \cdot (\lambda_{\text{max}}[K_S] + \sigma_S)^{-1} \\
    &=
    \frac{\lambda_{\text{min}}[K_S]}{[\lambda_{\text{min}}[K_S] + \sigma_S]} \cdot (\lambda_{\text{max}}[K_S] + \sigma_S)^{-1}.\label{eq: proof of cor1_3}
\end{align}
With Equation~\ref{eq: proof of cor1_1}, Equation~\ref{eq: proof of cor1_2} and Equation~\ref{eq: proof of cor1_3}, we can show that
\begin{align}
    \lambda_i (\hat\Theta_T(X_T, X_T)) 
    \geq
    \lambda_{\text{min}}(b Y_S^T Y_S b^T) 
    \cdot
    c_A
    \cdot
    [\frac{\lambda_{\text{min}}[K_S]}{\lambda_{\text{min}}[K_S] + \sigma_S}]
    \cdot
    \lambda_{\text{min}}[\hat\Theta_A(X_T, X_T)]
\end{align}
which is what we want.
\end{proof}

\subsection{Theorem 3 and Its Proof}\label{sec: Theorem 4 and Its Proof}
\begin{customthm}{3}\label{custhm: spectrum_B}
Suppose Assumption~\ref{cusassump: ntk} holds, the eigenvalue spectrum of the kernel matrix $\Theta_T^B(X_T, X_T)$ can be bounded as follows.
\begin{align}
    &\lambda_{i}(\Theta_B(X_T, X_T))
    \leq
    \lambda_{\text{max}}[k(a(X_T), X_S)k(X_S, a(X_T))]\nonumber\\
    &\cdot
    \lambda_{\text{max}}[[K_S + \sigma_S I]^{-2}]
    \cdot
    \lambda_{\text{max}}[Y_S Y_S^T]
\end{align}
and
\begin{align}
    &\lambda_{i}(\Theta_B(X_T, X_T))
    \geq
    \lambda_{\text{min}}[k(a(X_T), X_S)k(X_S, a(X_T))]\nonumber\\
    &\cdot
    \lambda_{\text{min}}[[K_S + \sigma_S I]^{-2}]
    \cdot
    \lambda_{\text{min}}[Y_S Y_S^T],
\end{align}
where $\lambda_i(\cdot)$ is the operator to output the $i$-th large eigenvalue.
\end{customthm}
\begin{proof}
Recall that $\Theta_T^B(x, x') = f_S(a(x))^T f_S(a(x'))$ and $f_S(\cdot) = k(\cdot, X_S)[K_S +\sigma_S I]^{-1} Y_S$. We can derive 
\begin{align}
    \Theta_T^B (X_T, X_T) = k(a(X_T), X_S) [K_S + \sigma_S I]^{-1} Y_S Y_S^T [K_S + \sigma_S I]^{-1} k(X_S, a(X_T))
\end{align}
Then, with similar process showned in the Appendix~\ref{sec: Theorem 3 and Its Proof}, we have
\begin{align}
    &\lambda_i(\Theta_T^B(X_T, X_T))
    \leq
    \lambda_{\max}(\Theta_T^B(X_T, X_T))\\
    &=
    \sup_{x\in\mathbb{R}^{c_T}|\|x\|_2 = 1}
    x^Tk(a(X_T), X_S) [K_S + \sigma_S I]^{-1} Y_S Y_S^T [K_S + \sigma_S I]^{-1} k(X_S, a(X_T))x\\
    &\leq
    \lambda_{\max}[Y_S Y_S^T]\cdot
    \sup_{x\in\mathbb{R}^{c_T}|\|x\|_2 = 1}
    x^Tk(a(X_T), X_S) [K_S + \sigma_S I]^{-2} k(X_S, a(X_T))x\\
    &\leq
    \lambda_{\max}[Y_S Y_S^T]\cdot
    \lambda_{\max}\{[K_S + \sigma_S I]^{-2}\}\cdot
    \sup_{x\in\mathbb{R}^{c_T}|\|x\|_2 = 1}
    x^Tk(a(X_T), X_S) k(X_S, a(X_T))x\\
    &=
    \lambda_{\max}[Y_S Y_S^T]\cdot
    \lambda_{\max}\{[K_S + \sigma_S I]^{-2}\}\cdot
    \lambda_{\max}[k(a(X_T), X_S)k(X_S, a(X_T))].
\end{align}

On the other hand, similarly, we can derive
\begin{align}
    &\lambda_i(\Theta_T^B(X_T, X_T))
    \geq
    \lambda_{\min}(\Theta_T^B(X_T, X_T))\\
    &=
    \inf_{x\in\mathbb{R}^{c_T}|\|x\|_2 = 1}
    x^Tk(a(X_T), X_S) [K_S + \sigma_S I]^{-1} Y_S Y_S^T [K_S + \sigma_S I]^{-1} k(X_S, a(X_T))x\\
    &\geq
    \lambda_{\min}[Y_S Y_S^T]\cdot
    \inf_{x\in\mathbb{R}^{c_T}|\|x\|_2 = 1}
    x^Tk(a(X_T), X_S) [K_S + \sigma_S I]^{-2} k(X_S, a(X_T))x\\
    &\geq
    \lambda_{\min}[Y_S Y_S^T]\cdot
    \lambda_{\min}\{[K_S + \sigma_S I]^{-2}\}\cdot
    \inf_{x\in\mathbb{R}^{c_T}|\|x\|_2 = 1}
    x^Tk(a(X_T), X_S) k(X_S, a(X_T))x\\
    &=
    \lambda_{\min}[Y_S Y_S^T]\cdot
    \lambda_{\min}\{[K_S + \sigma_S I]^{-2}\}\cdot
    \lambda_{\min}[k(a(X_T), X_S)k(X_S, a(X_T))]
\end{align}
which is what we desire.
\end{proof}

\subsection{Corollary 2 and Its Proof}\label{sec: Corollary 2 and Its Proof}
\begin{customassum}{3}\label{cusassump: cor_2}
Given a source dataset $D_S$ and a target and dataset $D_T$, there exists $c_B > 0$ such that
\begin{equation}
    \lambda_{\text{min}}[k(a(X_T), X_S)k(X_S, a(X_T))] 
    \geq
    c_B \cdot
    (\lambda_{\text{max}}[K_S])^2.
\end{equation}
\end{customassum}

\begin{customcor}{2}\label{cuscor: spectrum_B}
Suppose Assumption~\ref{cusassump: ntk}  and Assumption~\ref{cusassump: cor_2} hold, then we have
\begin{align}
    \lambda_i (\Theta_T^B(X_T, X_T))
    \geq
    c_B\cdot[\frac
    {\lambda_{\text{min}}[K_S]}{\lambda_{\text{min}}K_S + \sigma_S I}]^2
    \cdot
    \lambda_{\text{min}}[Y_S Y_S^T].
\end{align}
\end{customcor}
\begin{proof}
The proof for Corollary~\ref{cuscor: spectrum_B} is very similar to the proof for Corollary~\ref{cuscor: spectrum} which is shown in Appendix~\ref{sec: Corollary 1 and Its Proof}. First, we notice that
\begin{align}
    \lambda_{\min}\{[K_S + \sigma_S I]^{-2}\}
    &=
    \lambda_{\min}\{[\frac{K_S}{K_S + \sigma_S I}]^{2} \cdot K_S^{-2}\}\\
    &\geq
    \lambda_{\min}\{[\frac{K_S}{K_S + \sigma_S I}]^{2}\}
    \cdot\lambda_{\min}[K_S^{-2}]\\
    &=
    \lambda_{\min}\{[\frac{K_S}{K_S + \sigma_S I}]\}^{2}
    \cdot\lambda_{\max}[K_S]^{-2}\\
    &=
    [\frac{\lambda_{\min}[K_S]}{\lambda_{\min}[K_S + \sigma_S I]}]^{2}
    \cdot\lambda_{\max}[K_S]^{-2}
    \label{proof: cor_2_1}
\end{align}
Then, combining Theorem~\ref{custhm: spectrum_B} and Eq.~\ref{proof: cor_2_1}, we can conclude that
\begin{equation}
     \lambda_i (\Theta_T^B(X_T, X_T))
    \geq
    c_B\cdot[\frac
    {\lambda_{\text{min}}[K_S]}{\lambda_{\text{min}}K_S + \sigma_S I}]^2
    \cdot
    \lambda_{\text{min}}[Y_S Y_S^T].
\end{equation}
which is what we want.
\end{proof}

\subsection{Theorem 4 and Its Proof}\label{sec: Theorem 2 and Its Proof}
\begin{customassum}{4}\label{cusassump: general} 
To derive an upper bound of the generalization gap, we consider the following assumptions:
\begin{enumerate}
        \item $\|(x, y) - (x', y')\|_2 \leq\Gamma_{\mathfrak{D}_T},~\forall (x, y), (x', y')\sim\mathfrak{D}_T$,
        \item $\|\nabla_{(x, y)} g(x, y)\|_2 \leq L_{\mathfrak{D}_T},~\forall(x, y)\sim\mathfrak{D}_T$,
        \item $\|\nabla_{f} g(x, y)\|\leq\rho,~\forall      f\in\mathcal{H}_{\Theta_T}^{c_T}$,
        \item $\frac{T\cdot\rho^2}{N_T^2}\sum_{(x_t, y_t)\sim D_T \atop ({x'}_t, {y'}_t)\sim D_T} |\Theta_T (x_t, x_t')| \leq B^2$.
\end{enumerate}
\end{customassum}
\begin{customthm}{4}[Upper bound of generalization gap]\label{custhm:gen}
Given a $N_T$-sample target dataset $D_T$, sampled from the target distribution $\mathfrak{D}_T$. Suppose that Assumption~\ref{assump: general} holds, then the generalization gap defined in Equation~\ref{eq: framework} can be bounded as follows with probability at least $1-\delta$:
\begin{align}
    &\mathbb{E}_{(x_t,y_t)\sim\mathfrak{D}_T}[g(x_t, y_t)] - \sum_{(x_i, y_i)\in D_T} \frac{g(x_i, y_i)}{N_T} \nonumber\\
    &\leq 
    \frac{2 \rho B \sqrt{T}}{N_T} 
    \sum_{(x_t, y_t)\in D_T \atop (x_t', y_t')\in D_T} |\Theta_T (x_t, x'_y)|
    + 3 L_{\mathfrak{D}_T} \Gamma_{\mathfrak{D}_T} \sqrt{\frac{\log{\frac{2}{\delta}}}{2N_T}}\label{eq:cusgengap},~\forall g\in\mathcal{G}.
\end{align}
\end{customthm}
\begin{proof}
Theorem~\ref{custhm:gen} is the composition of Theorem 3 in \citep{chung2024rethinking} and Equation 4 in \citep{chen2024analyzing}. Under the first item to the third item of Assumption~\ref{cusassump: general}, \citet{chung2024rethinking} have shown that the generalization gap can be bounded as 
\begin{align}
    &\mathbb{E}_{(x_t,y_t)\sim\mathfrak{D}_T}[g(x_t, y_t)] - \sum_{(x_i, y_i)\in D_T} \frac{g(x_i, y_i)}{N_T} \nonumber\\
    &\leq 
    2\Hat{\mathfrak{R}}_{D_T}(\mathcal{G})
    + 3 L_{\mathfrak{D}_T} \Gamma_{\mathfrak{D}_T} \sqrt{\frac{\log{\frac{2}{\delta}}}{2N_T}},~\forall g\in\mathcal{G}.\label{eq: proof of thm2_1}
\end{align}
where $\Hat{\mathfrak{R}}_{D_T}(\mathcal{G})$ is the empirical Rademacher's complexity~\citep{Mohri}. Then, with the fourth item of Assumption~\ref{cusassump: general}, \citet{chen2024analyzing} have proven that the empirical Rademacher's complexity can be characterized by the kernel matrix $\Theta_T (X_T, X_T)$. Specifically,
\begin{equation}
    \Hat{\mathfrak{R}}_{D_T}(\mathcal{G}) \leq \frac{\rho B \sqrt{T}}{N_T} 
    \sum_{(x_t, y_t)\in D_T \atop (x_t', y_t')\in D_T} |\Theta_T (x_t, x'_y)|
    \label{eq: proof of thm2_2}
\end{equation}
Combining Equation~\ref{eq: proof of thm2_1} and Equation~\ref{eq: proof of thm2_2}, we can derive Equation~\ref{eq:cusgengap}, which is what we want.
\end{proof}

\clearpage

\section{Experimental Details of Main Results}\label{sec: exp details}
In this section, we present our experimental settings in detail.


\begin{figure*}[ht]
    \centering
    \includegraphics[width=0.85\textwidth]{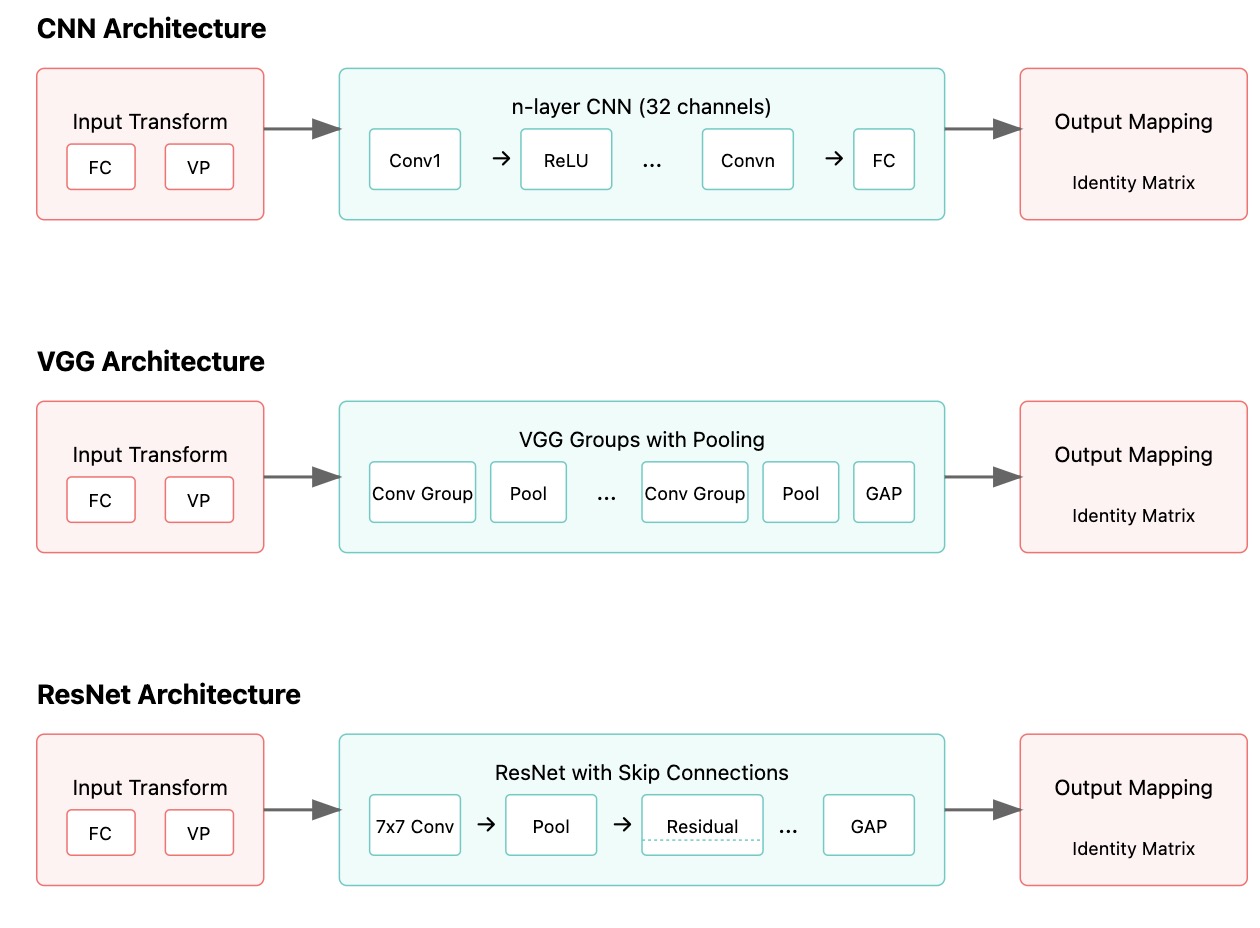}
    \caption{Model Reprogramming architecture variations with different backbone networks. Each architecture consists of three key components: (1) An input transformation layer that can be implemented as either a fully connected network (FC) or visual prompt (VP), (2) A frozen backbone network that varies between CNN (simple sequential layers), VGG (grouped convolutions with pooling), or ResNet (residual blocks with skip connections), and (3) An identity matrix as the output mapping layer.}
    \label{fig:mr-architectures}
\end{figure*}

\subsection{Source Model Structure}\label{sec: exp source model}
We chose three different models as the structures of the source model to perform our experiments:
\begin{itemize}
    \item \textbf{Convolutional Neural Networks (CNNs)}: The $n$-layer CNNs discussed in this paper include only the $n$ convolutional layers and do not count the final fully connected layer. All convolutional layers, except the first one, have 32 input and output channels, with ReLU used as the activation function. No pooling layers are used in this network.

    \item \textbf{Visual Geometry Group (VGG)}: The VGG network consists of n convolutional groups, each containing a specified number of convolutional layers with shared channel width. Each group is followed by average pooling, with channel width doubling between groups. The network concludes with global average pooling and classification. When configured with a single group, this architecture reduces to the CNN implementation described above.

    \item \textbf{Residual Neural Network (ResNet)} The ResNet begins with a strided 7×7 convolution and pooling layer, followed by n groups of residual blocks. Each residual block contains two 3×3 convolutions with skip connections implemented through identity mappings or 1×1 convolutions for channel matching. Groups are separated by average pooling layers with channel width doubling between groups. The network terminates with global average pooling and classification.
\end{itemize}

\section{Experimental Details of Large-scale Model}\label{sec: experiments}
\subsection{Experimental Setup}\label{sec: exp setup}
To complement our main results, we conducted additional experiments using the CLIP (ResNet-50) architecture to validate our theoretical findings at scale. Our experimental framework incorporates several  modifications to the traditional model reprogramming setup. While conventional approaches utilize a single CNN as the source model with input transformation and output mapping layers, we propose a more sophisticated framework: we leverage CLIP as a fixed feature extractor, followed by a trainable CNN of varying depths. This modification allows us to systematically investigate how network depth affects model reprogramming performance while maintaining the benefits of CLIP's pre-trained representations.

\paragraph{Source Model Structure} Our source model architecture consists of two key components working in tandem: A CLIP model (ResNet-50) that serves as a fixed feature extractor, leveraging its pre-trained weights to provide rich semantic representations. A trainable CNN component that processes CLIP's feature outputs. This CNN consists of n sequential convolutional layers (where n varies from 1 to 6 in our experiments), each maintaining 32 input and output channels with ReLU activation. During source training, only these CNN parameters are optimized while CLIP's weights remain frozen.

\paragraph{Target Model Adaptation} For target domain adaptation, we augment the trained source model (CLIP+CNN) with an input transformation layer and an output mapping. During adaptation, both CLIP and CNN weights remain fixed, with only the input transformation and output mapping being optimized. We examine two types of input transformation structures: FC and VP.

\paragraph{Datasets} Our experiments utilize ImageNet-10 as the source dataset, with CIFAR-10 and SVHN as target datasets. All images are resized to 32×32.

\paragraph{Model Details} Each CNN layer maintains 32 input and output channels with ReLU activation, excluding the final fully connected layer. The output mapping is implemented as a linear matrix.

\subsection{Results}\label{sec: results}
Tables 1-4 present our experimental results across all dataset combinations. Several key observations emerge:

\paragraph{Depth Impact} As predicted by our theory, increasing network depth from 1 to 6 layers consistently improves both source and target performance. For example, in CIFAR10 → ImageNet10 with FC structure, source accuracy improves from 58.20

\paragraph{Cross-Domain Transfer} The effectiveness of transfer learning varies across dataset pairs. VP structure demonstrates superior performance compared to FC, particularly in deeper networks. In CIFAR10 → SVHN scenario, VP achieves 44.68

\begin{figure}[H]
    \centering
    \centerline{\includegraphics[width=0.25\linewidth]{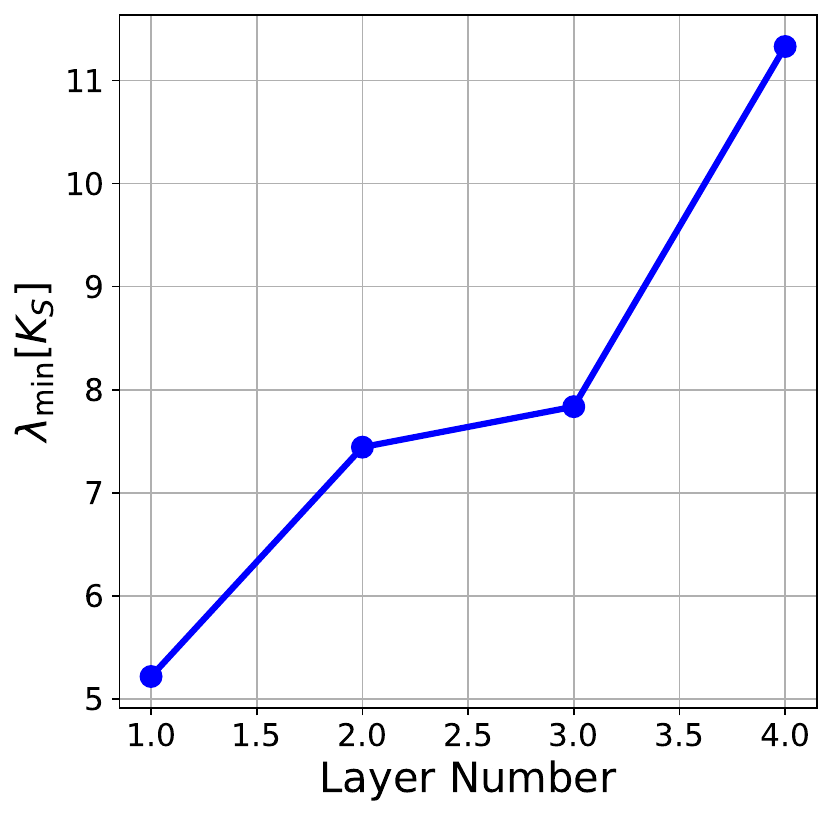}}
    \vspace{-10pt}
    \caption{$\lambda_{\min}[K_S]$ v.s. source model's depth (CLIP)}
    \label{fig: NTK_spectrum_clip}
    \vspace{-10pt}
\end{figure}

\paragraph{Eigenvalue Spectrum of NTK}
We compute the eigenvalue spectrum of the CLIP model here. The experiment is recorded in the Fig~\ref{fig: NTK_spectrum_clip}. The experimental results suggest that the $\lambda_{\min}[K_S]$ increase as we deepen the CLIP. Based on Corollary~\ref{cor: spectrum} and Corollary~\ref{cor: spectrum_B}, we can infer that both target loss and source loss would decrease as depths of CLIP increase.

\paragraph{Source Loss v.s. Target Loss} The source loss and target loss are recorded in the Table~\ref{Tab: CLIP}. We observe that both source loss and target loss decrease as we deepen the source model, and strong correlation between source and target losses across different depths, supporting our theoretical prediction that target model success depends on source model performance. This correlation is particularly evident in CE loss experiments.

\begin{figure}[t]
\centering
\begin{minipage}{\textwidth}
\includegraphics[width=\textwidth]{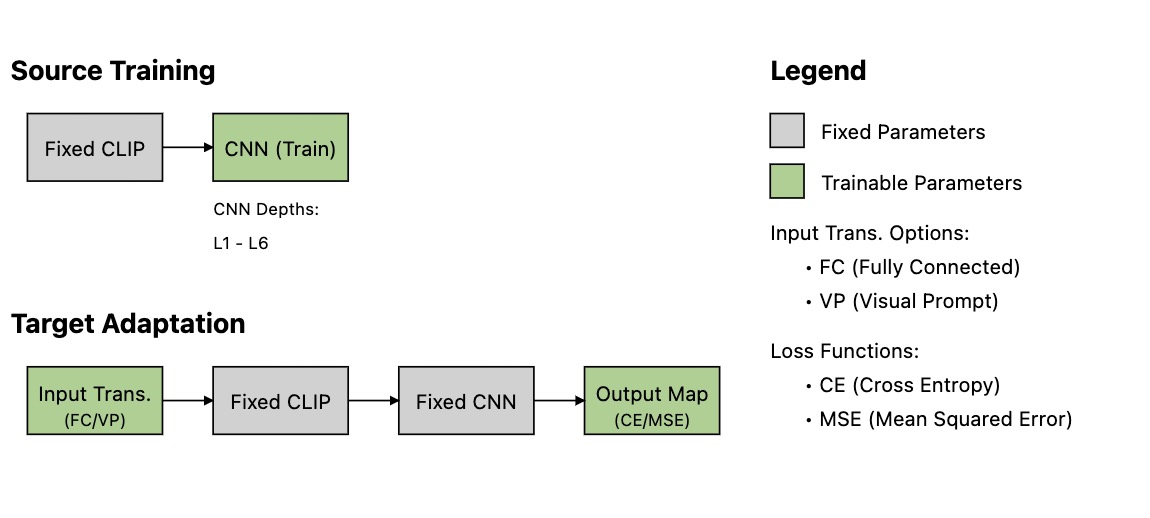}
\caption{Overview of our model reprogramming framework. \textbf{Top:} Source training process with fixed CLIP and trainable CNN of varying depths (L1-L6). \textbf{Bottom:} Target adaptation process with trainable input transformation (FC/VP) and output mapping (CE/MSE), while keeping CLIP and CNN fixed. Gray blocks indicate fixed parameters, while green blocks indicate trainable parameters.}
\label{fig:model_architecture}
\end{minipage}
\end{figure}
As illustrated in Figure~\ref{fig:model_architecture}, our framework consists of two main stages: source training and target adaptation. During source training, we fix the CLIP model and only train the CNN with varying depths. For target adaptation, we augment the fixed source model (CLIP+CNN) with trainable input transformation and output mapping components.

These results validate our theoretical framework linking eigenvalue spectrum to model performance, while also providing practical insights for model reprogramming applications.

\begin{figure}[H]
	\centering
	\begin{minipage}{0.24\linewidth}
		\centerline{\includegraphics[width=\textwidth]{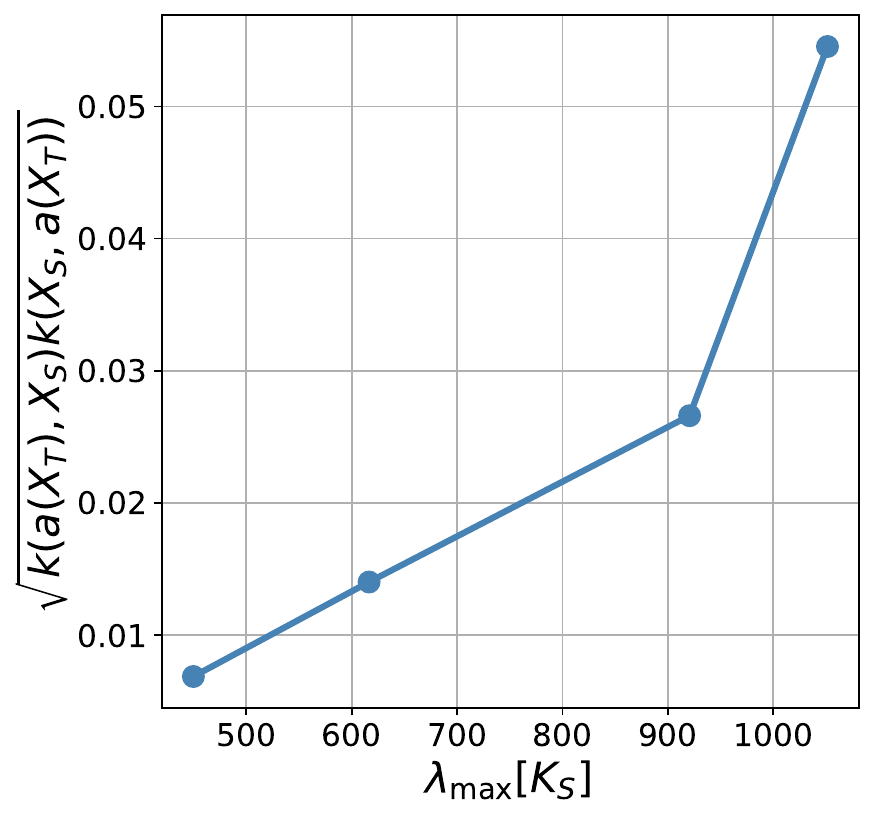}}
         \centerline{\small(a)ImageNet10$\rightarrow$ CIFAR10 (FC)}
	\end{minipage}
    \hspace{0.06cm}
	\begin{minipage}{0.24\linewidth}
		\centerline{\includegraphics[width=\textwidth]{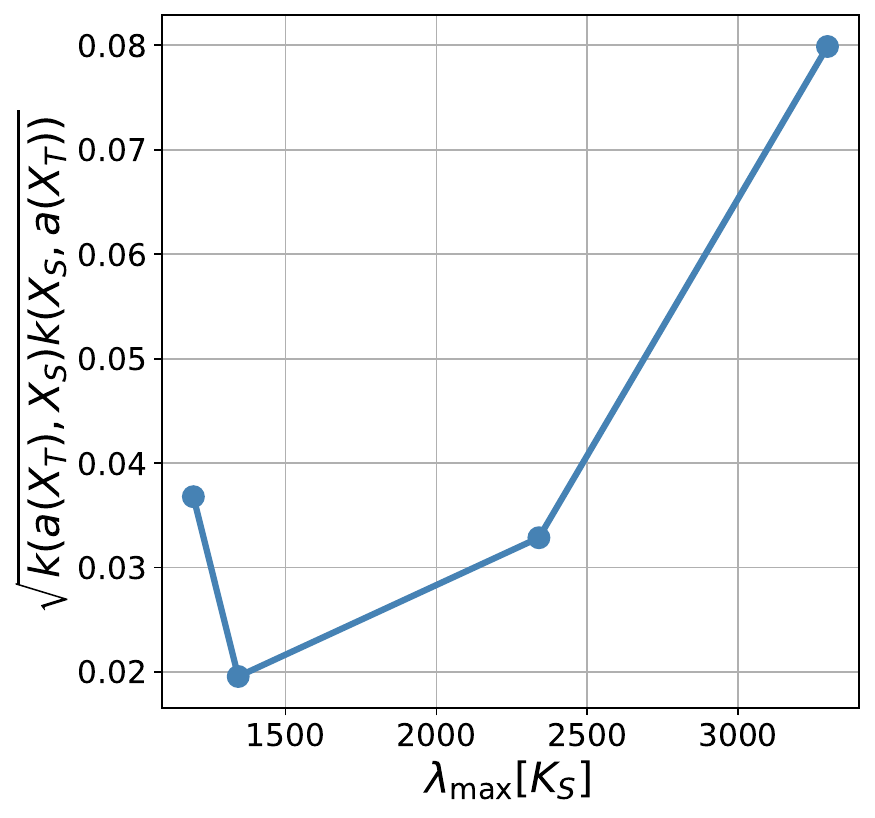}}
         \centerline{\small(b)ImageNet10$\rightarrow$ CIFAR10 (VP) }
	\end{minipage}
	\begin{minipage}{0.24\linewidth}
		\centerline{\includegraphics[width=\textwidth]{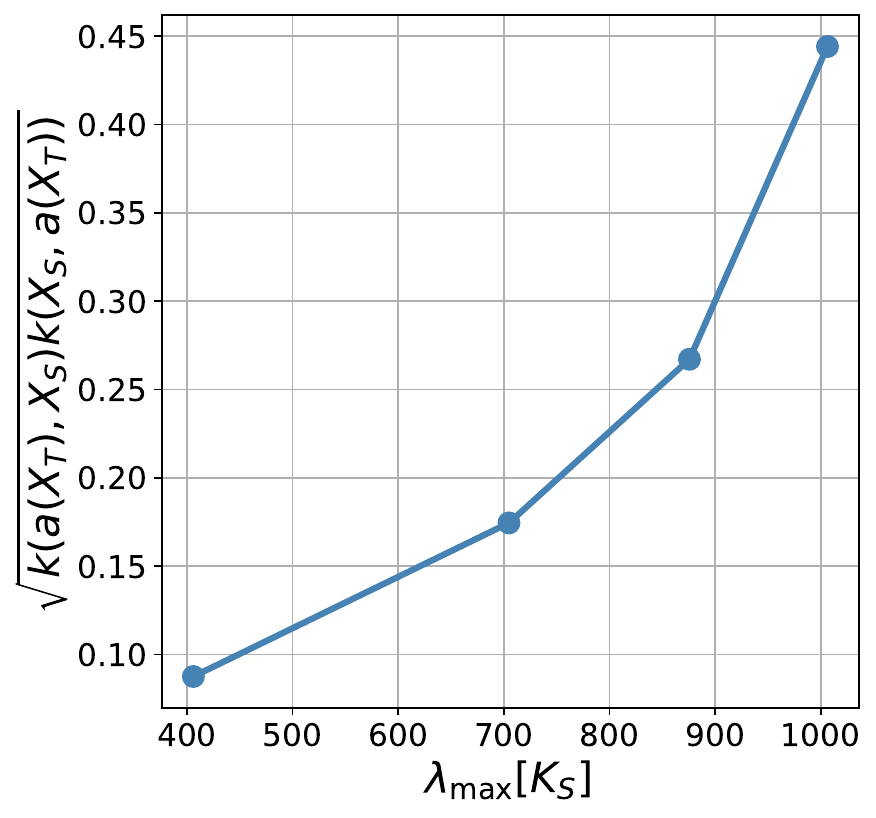}}
         \centerline{\small(c) ImageNet10$\rightarrow$ SVHN (FC)}
	\end{minipage}
	\begin{minipage}{0.24\linewidth}
		\centerline{\includegraphics[width=\textwidth]{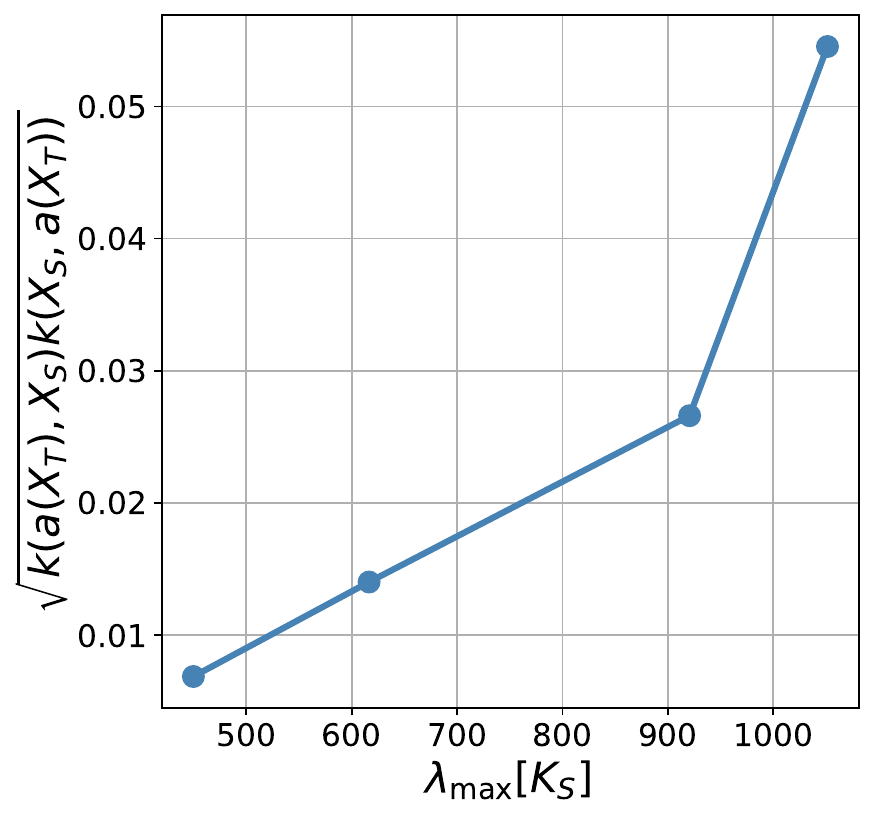}}
         \centerline{\small(d) ImageNet10$\rightarrow$ CIFAR10 (VP)}
	\end{minipage}
	\caption{Assumption Justification} 
	\label{fig: assumption justification_CLIP}
\end{figure}

\paragraph{Assumption Justification}
At the end of this section, we conducted experiments to verify Assumption~\ref{assump: cor_2}. The experimental result is recorded in the Fig.~\ref{fig: assumption justification_CLIP}. In this figure, we notice that $\sqrt{\lambda_{\min}[k(a(X_T), X_S)k(X_S, a(X_T))]}$ grows much faster than $\lambda_{\max}[K_S]$ (higher than first order), which manifests that our assumption is valid.

\begin{table}[t]
\caption{Performance comparison with ImageNet10 as source data, showing VP source and both FC/VP target results. Each cell shows accuracy / loss.}
\centering
\setlength{\tabcolsep}{4pt}
\small
\begin{tabular}{c|c|c|c}
\toprule[1pt]
\multicolumn{4}{c}{ImageNet10 $\rightarrow$ CIFAR10 (Cross-Entropy)} \\
\midrule
Layer & Source (VP) & Target (FC) & Target (VP) \\
\midrule
1 & 52.42 / 1.364 & 48.50 / 1.412 & 46.63 / 1.490 \\
2 & 60.12 / 1.196 & 48.53 / 1.432 & 50.07 / 1.401 \\
3 & 62.27 / 1.127 & 50.04 / 1.401 & 49.52 / 1.399 \\
4 & 65.35 / 1.024 & 51.50 / 1.353 & 47.95 / 1.429 \\
\bottomrule[1pt]
\end{tabular}

\vspace{0.3cm}

\begin{tabular}{c|c|c|c}
\toprule[1pt]
\multicolumn{4}{c}{ImageNet10 $\rightarrow$ SVHN (Cross-Entropy)} \\
\midrule
Layer & Source (VP) & Target (FC) & Target (VP) \\
\midrule
1 & 45.15 / 1.565 & 30.08 / 2.036 & 28.65 / 2.049 \\
2 & 48.08 / 1.450 & 29.15 / 2.034 & 32.96 / 1.941 \\
3 & 50.88 / 1.404 & 34.81 / 1.891 & 34.86 / 1.883 \\
4 & 55.35 / 1.292 & 35.64 / 1.876 & 32.92 / 1.937 \\
\bottomrule[1pt]
\end{tabular}




\label{Tab: CLIP}
\end{table}

\end{document}